\documentclass[final,3p,times,onecolumn]{elsarticle}
\pdfoutput=1

\usepackage{subfig}
\usepackage{amsmath}
\usepackage{upgreek}
\usepackage{multirow}
\usepackage{siunitx}
\usepackage[export]{adjustbox}
\usepackage[linesnumbered,ruled, noend]{algorithm2e}
\usepackage{enumitem}
\usepackage{booktabs}
\usepackage{comment}
\usepackage{svg}
\usepackage{scalerel}
\usepackage{todonotes}

\usepackage[hidelinks]{hyperref}
\usepackage[nameinlink, capitalise]{cleveref}

\Crefname{figure}{Fig.}{Figs.}
\Crefname{equation}{Eq.}{Eqs.}

\let\today\relax
\makeatletter
\def\ps@pprintTitle{%
    \let\@oddhead\@empty
    \let\@evenhead\@empty
    \def\@oddfoot{\footnotesize\itshape
         {Preprint} \hfill\today}%
    \let\@evenfoot\@oddfoot
    }
\makeatother

\journal{}
\begin{document}

\begin{frontmatter}

\title{On architectural choices for interpretability and thermodynamic consistency in Physically Recurrent Neural Networks in the low-data regime}
\author[a]{M. A. Maia}
\author[b]{K. A. Meyer}
\author[a]{A. M. C. M. van Gils}
\author[a]{I. B. C. M. Rocha}
\author[a]{F. P. van der Meer}

\address[a]{{Delft University of Technology, Department of Civil Engineering and Geosciences}, {P.O.Box 5048}, 2600GA, Delft, {The Netherlands}}
\address[b]{Chalmers University of Technology, Department of Mechanical Engineering, Gothenburg, Sweden}

\begin{abstract}
In this paper, we unravel the effect of different decoder architectures on the interpretability of the latent space of the Physically Recurrent Neural Network. Particular emphasis is given to a new weight normalization constraint, which acts as a regularization technique and enables robust training in the low-data regime. A brief visual exploration illustrates how these changes impact the latent space and how the fictitious stress can align with the true state of the RVE without explicit training. Reaping the benefits of a meaningful latent space, a case study illustrates how information from the microscopic level can be retrieved and incorporated into a multi-task approach that does not require extra parameters or larger training sets. Another key contribution shows that a specific architectural choice can naturally lead to a thermodynamically consistent formulation. By enforcing an adjoint encoder–decoder structure with positive scalar contributions, this modification ensures energy consistency across scales and non-negative dissipation, leading to even lower training requirements. This alternative completes the study on interpretability, inductive bias, and thermodynamic consistency, and  demonstrates that data efficiency can be improved with careful architectural choices rooted in the underlying physics.
\end{abstract}


\end{frontmatter}


\newcommand{\Densedec}{\textit{Dense}}
\newcommand{\Componentdec}{\textit{Component}}
\newcommand{\Pointdec}{\textit{Point}}
\newcommand{\TCdec}{\textit{TC}}
\newcommand{\Normdensedec}{\textit{Normalized dense}}
\newcommand{\Normcomponentdec}{\textit{Normalized component}}
\newcommand{\Normpointdec}{\textit{Normalized point}}
\newcommand{\NormTCdec}{\textit{Normalized TC}}
\newcommand{\fesqr}{FE$^2$}
\newcommand{\pd}[2]{\frac{\partial{#1}}{\partial{#2}}}
\newcommand{\strainvec}{\boldsymbol{\varepsilon}}
\newcommand{\stressrot}{\boldsymbol{\sigma}_{\textrm{F}}}
\newcommand{\predstresslocal}{\widehat{\boldsymbol{\sigma}}^{\Omega}_{\textrm{L}}}
\newcommand{\predstressrot}{\widehat{\boldsymbol{\sigma}}^{\Omega}_{\textrm{F}}}
\newcommand{\predstress}{\widehat{\boldsymbol{\sigma}}^{\Omega}}
\newcommand{\strainstresscurve}{\boldsymbol{\varepsilon}\text{-}\boldsymbol{\sigma}}
\newcommand{\intvarvec}{\boldsymbol{\alpha}}
\newcommand{\macrostrainvec}{\boldsymbol{\varepsilon}^{\Omega}}
\newcommand{\dotmacrostrainvec}{\dot{\boldsymbol{\varepsilon}}^{\Omega}}
\newcommand{\macrostressvec}{\boldsymbol{\sigma}^{\Omega}}
\newcommand{\microstressvec}{\boldsymbol{\sigma}^{\omega}}
\newcommand{\microstrainvec}{\boldsymbol{\varepsilon}^{\omega}}
\newcommand{\macrodispfield}{\mathbf{u}^\Omega}
\newcommand{\microdispfield}{\mathbf{u}^\omega}
\newcommand{\dispfield}{\mathbf{u}}
\newcommand{\weights}{\mathbf{W}}
\newcommand{\Wjmat}{\mathbf{W}_{j}}
\newcommand{\Ljmat}{\mathbf{W}_j}
\newcommand{\biases}{\mathbf{b}}
\newcommand{\hidden}{\mathbf{h}}
\newcommand{\histunits}{\mathbf{h}}
\newcommand{\stiffmatrix}{\mathbf{D}}
\newcommand{\xvec}{\mathbf{x}}
\newcommand{\Bmat}{\mathbf{B}}
\newcommand{\macrodmat}{\mathbf{D}^{\Omega}}
\newcommand{\Idtensor}{\mathbf{I}}
\newcommand{\stress}[2][]{\boldsymbol{\sigma}_{#1}^{#2}}
\newcommand{\intvar}[2][]{\boldsymbol{\alpha}_{#1}^{#2}}
\newcommand{\predmacrostress}{\widehat{\boldsymbol{\sigma}}^{\Omega}}
\newcommand{\norm}[1]{\left\lVert#1\right\rVert}
\newcommand{\weightcon}{w_{{\widehat{\sigma}^{\Omega}_{i}}-\sigma_{k,j}}}
\newcommand{\testunlgp}{\mathcal{T}_{\textrm{GP}}}

\section{Introduction}
\label{sec:intro}

In many engineering fields, surrogate models have been key in enabling otherwise prohibitive or time-consuming simulations. In the material modeling community, these models have been employed in applications such as material characterization, inverse parameter identification, structural optimization, sensitivity analysis, real-time digital twins, and more. A particularly active area is their use to accelerate concurrent multiscale simulations (FE$^2$). In FE$^2$, two scales (macro and micro) exchange information and are solved iteratively. The main difference from regular Finite Element (FE) is that no explicit constitutive model relating strains to stresses at the macroscale exists. Instead, a micromodel consisting of another FE mesh representing the geometry of the material is embedded at each integration point of the macroscale mesh, and its homogenized response is used to describe the material at the upper scale. A full-order multiscale simulation, however, requires solving numerous local finite element (FE) problems at each time increment, quickly becoming intractable for large-scale and time-dependent analyses. 

Therefore, replacing this rich but computationally intensive representation of a heterogeneous material can yield substantial gains. However, this usually comes at the expense of generalization beyond the training data and interpretability. In the most straightforward approach, the problem amounts to learning the macroscopic response from pre-computed full-order micromodel analyses, often relying on pairs of homogenized strain-stress and treating the FE$^2$ solver as a black-box model. This approach has been successfully applied to a range of homogenization problems. However, the shortcomings of purely data-driven models become particularly evident when history dependence is factored in. In this case, surrogate models cannot predict the macroscopic response from current strains alone, pushing the community toward models with some form of memory. For example, with solutions based on neural networks (NNs), this issue is typically addressed through recurrent architectures. 

Compared to feed-forward NNs, Recurrent NNs (RNNs) have additional parameters that carry information from one time step to another, making them a compelling choice for approximating history-dependent materials. The idea of a persisting state that evolves in time parallels the evolution of internal variables in a micromodel. Still, the interpretation of this so-called \textit{hidden state} is not as straightforward as the micromodel's state. In more advanced recurrent architectures, such as the Gated Recurrent Unit (GRU) and Long Short-Term Memory (LSTM), the choice between discarding or incorporating information throughout time is managed by mechanisms called gates (\textit{e.g.} input, forget, update, reset, and output gates). These networks capture long-term dependencies better and struggle less with vanishing or exploding gradients. In GRUs, the hidden state serves as both \textit{memory} and \textit{output}, in LSTMs, the architecture handles these as two separate parts. To do this, LSTMs implement a dedicated additional set of variables known as the \textit{cell state}, a long-term conveyor belt that stores information in long sequences more reliably and passes only selected parts of it to the output. 

When dealing with mixed types of features, namely temporal and non-temporal, one way to preserve accuracy in recurrent models is to tailor the memory mechanisms. For example, to incorporate microstructural descriptors in GRUs alongside the typical strain-stress pair as input, \cite{Mozaffar26414} proposed a separate information flow to handle non-temporal data and prevent them from being corrupted over time. \cite{Guo2021cpi} followed the same philosophy to combine LSTMs and CNNs for the identification of the properties of elastoplastic materials and to process sequential full-field displacement data. Despite these successful targeted modifications in both works, the relationship between the material's internal variables and the network's remains unexplored. Typically, the interpretability aspect is explored in the context of homogeneous materials \cite{Koeppe2021, BonattiMohr2021}. 

Another interesting approach is the Minimum State Cells (MSC) \cite{BonattiMohr2021}. The authors designed a special RNN with the idea that a universal material model should feature a small, physically meaningful number of state variables, rather than relying on excessive memory units. The arbitrarily large number of internal variables in NNs hinders the identification of interpretable state variables, a manifestation of the so-called \textit{black-box} nature of data-driven methods, and can lead to overfitting and poor extrapolation. Examples in \cite{BonattiMohr2021} showed a good correlation between the internal variables of the network and those of the constitutive model of reference. However, the novel framework, as well as the GRU-based methods discussed previously, fail to extrapolate to sequences longer than those seen during training, and, most importantly, do not satisfy fundamental material properties such as stationarity and self-consistency. 

By stationarity, we mean that the internal variables ($\boldsymbol{\alpha}$) and stresses ($\boldsymbol{\sigma}$) of rate-independent and path-dependent models remain unchanged when applying a zero-strain increment ($\Delta\boldsymbol{\varepsilon}=\mathbf{0}\rightarrow\Delta\boldsymbol{\alpha} = \mathbf{0}, \Delta\boldsymbol{\sigma}=\mathbf{0}$). Self-consistency refers to the constitutive model's ability to predict the same state regardless of the loading-path discretization. Several approaches emulate these properties in NN-based solutions, from soft constraints to hard-wiring the mathematical properties within their architectures. \cite{Rosenkranz2023} further discusses this, presenting a comparative study based on three basic groups: black-box NNs, NNs enforcing physics in a weak form, and NNs enforcing physics in a strong form. The latter hard-code physics via custom architectures to ensure that the constraint is fulfilled in any loading condition. In this new paradigm, Input Convex Neural Networks (ICNNs) and automatic differentiation are key tools, often employed in energy-based formulations (instead of the usual strain-stress direct mapping) from which stresses and other quantities can be derived \cite{VLASSIS2021,HE2022,EGHBALIAN2023}. Another example in this group is \cite{HEIDENREICH2025}, where the authors modify the update gate of a GRU cell to satisfy the first property automatically and the second for sufficiently small increments. They followed the MSC philosophy of \cite{BonattiMohr2021}, but did not compare the network's internal variables with those of the micromodel. 

Although these works focus on enforcing certain physical properties in the NN, the related question of whether the learned state variables \textit{can} be given a meaningful physical interpretation is gaining traction. In \cite{MASI2026}, inelastic constitutive models are approximated under hard thermodynamic constraints from stress-strain data, and the NN learns the internal state variables. In this case, despite these variables not being uniquely identifiable from macroscopic data and, therefore, not being interpretable as unique physical internal variables, they provide useful insights, absent in the monotonic training data, into the path-dependent nature of the material.  

Similarly, in \cite{FLASCHEL2025}, the authors compared the evolution of the internal variables learned within a convex NN architecture with the corresponding variables of a reference constitutive model. In that work, two thermodynamic potentials are learned with thermodynamic consistency enforced by design. Again, different choices of internal variables can result in the same material response. However, their results show that the learned variables can recover the reference variables through a linear mapping. Building on these observations, \cite{RAJETAL2026} further investigates the identifiability of learned internal variables. The authors show that, under controllability assumptions, the internal variables of two distinguished models \textit{are} identifiable up to a linear transformation. These works reflect a shift toward understanding what NN-based models learn beyond their predictive accuracy, and a growing concern for assessing the physical meaning, reliability, and trustworthiness of their latent representations. 

Another important contribution in this field is \cite{JI2026}. The authors introduce Inelastic Constitutive Kolmogorov-Arnold Networks (ICKANs) for the automated discovery of interpretable inelastic constitutive models. Their approach offers another example of how NN structure can be designed to yield interpretable material models. In this contribution, our perspective on interpretable architectures is aligned most closely with \citet{SHIN2023}. In that work, a visualization of the Deep Material Network (DMN) was proposed as an analogous unit cell, where the network parameters are interpreted in terms of material fractions and orientations, providing an intuitive representation of the learned homogenization structure. The authors further exploit this representation in a recursive quilting strategy to initialize deeper networks and improve their accuracy and calibration. Most importantly, their network can reproduce both the macroscopic homogenized response and, in some cases, statistical features of the local response, illustrating how a network architecture can retain a direct connection to the underlying micromechanics. 

In \cite{Maiaetal2023}, we propose a hybrid model, the Physically Recurrent Neural Network (PRNN), to predict the macroscopic stress of heterogeneous and history-dependent materials. The core idea is to combine an encoder-decoder architecture with intact embedded constitutive models. The encoder learns a set of values, interpreted as strains, that are fed to (fictitious) material points associated with the same constitutive models as in the micromodel. These points, in turn, compute the stress and, for a history-dependent model, the updated internal variables. The collection of fictitious stresses is passed to the decoder to obtain the macroscopic stresses. At the same time, the internal variables are automatically updated according to the physics postulated in the constitutive model, effectively serving as the memory of the network. In contrast to fully data-driven architectures, the quantities passed to and from the embedded material models, which we refer to as a \textit{material layer}, have a prescribed physical interpretation. Furthermore, the material models and properties are directly linked to those of the underlying micromodel. 

This hybrid design allows PRNNs to generalize well with small training datasets \cite{Maiaetal2023,Maiaetal2024} and facilitated their extension to different multiscale problems, including microscale debonding \cite{kovacs2024}, uncertainty quantification \cite{kovacs2025}, and hierarchical modeling of woven composites \cite{ghane2025woven}. A common thread in these applications is that the material layer is leveraged for its physical interpretation (e.g. by exploring the fact that the material properties in the embedded models can be modified directly in \cite{kovacs2025}). Based on these developments, we now turn our attention to how architectural choices in the PRNN affect the interpretability of this latent space and the data efficiency of the model. In particular, we focus on the decoder, which determines how the fictitious stresses are combined to recover the macroscopic response. Although arbitrary multilayer perceptrons can represent both the encoder and decoder, our study restricts the decoder to a linear form and investigates different architectural constraints, with particular emphasis on a new weight-normalization constraint. This constraint acts as a form of regularization and enables robust training in the low-data regime. We also embed only elastoplastic models in the material layer. This choice simplifies model selection and makes it easier to visualize and interpret the impact of these changes. 

The low-data regime is especially relevant for computational homogenization, since generating sufficiently large training datasets can itself become a computational bottleneck. Data efficiency is also important when surrogate models are trained from experimental data, where additional testing means time-consuming and costly campaigns. Furthermore, models that require only a small amount of data are better suited to settings in which the surrogate is updated as additional information about the micromodel becomes available. This provides not only a practical motivation but also a more severe test of the network's generalization capabilities. Data augmentation \cite{CHEUNG2024b} and the combination of data from different fidelity levels \cite{GHANE2025} can reduce the cost of generating training data, but intrinsically data-efficient models remain necessary.

We closely examine the relationship between fictitious quantities and their microscopic counterparts, and through a visual analysis, we show that specific architectural choices can produce a meaningful alignment between the fictitious stress and the true microscopic state of the micromodel, even though these quantities are not used for training. We then exploit this finding to retrieve microscopic information from the surrogate and incorporate it into a multi-task formulation without introducing additional parameters or requiring larger training datasets.

Finally, we investigate how architectural choices can also encode thermodynamic consistency into the network. By enforcing an adjoint relationship between the encoder and decoder and restricting their scalar contributions to be positive, the resulting architecture satisfies the Hill-Mandel energetic consistency condition between the microscopic and macroscopic scales, and obeys the non-negative dissipation constraint. All of these developments and findings show how relatively simple architectural choices can improve data efficiency, provide insight into the learned latent space, and enable the recovery of microscopic information.

In the following, we present a short summary of the main aspects involved in computational homogenization and where exactly the PRNN would fit in it, including a brief description of its main elements. Then, the different decoder connectivities are discussed in \Cref{sec:decoder}, followed by an accuracy assessment in the low-data regime in \Cref{sec:accuracylowdata}. In \Cref{sec:latent}, we illustrate the link between the latent space of the PRNN and the microscopic quantities, and in \Cref{sec:multitask}, we use a case study to illustrate to what extent macroscopic and microscopic quantities can be learned simultaneously without compromising performance. Finally, the main conclusions of this work are summarized in \Cref{sec:conclusion}.

\section{Computational homogenization}
\label{sec:homogenization}

In this section, we present a brief overview of the two approaches to concurrent multiscale modeling explored in this work. One based on the full-order micromodel solution and the other based on the use of a surrogate model to replace the micromodel in the multiscale problem, specifically a Physically Recurrent Neural Network (PRNN).  

\subsection{Concurrent multiscale model}
\label{subsec:micro}

\begin{figure}[!ht]
\centering
 \includegraphics[width=.7\textwidth]{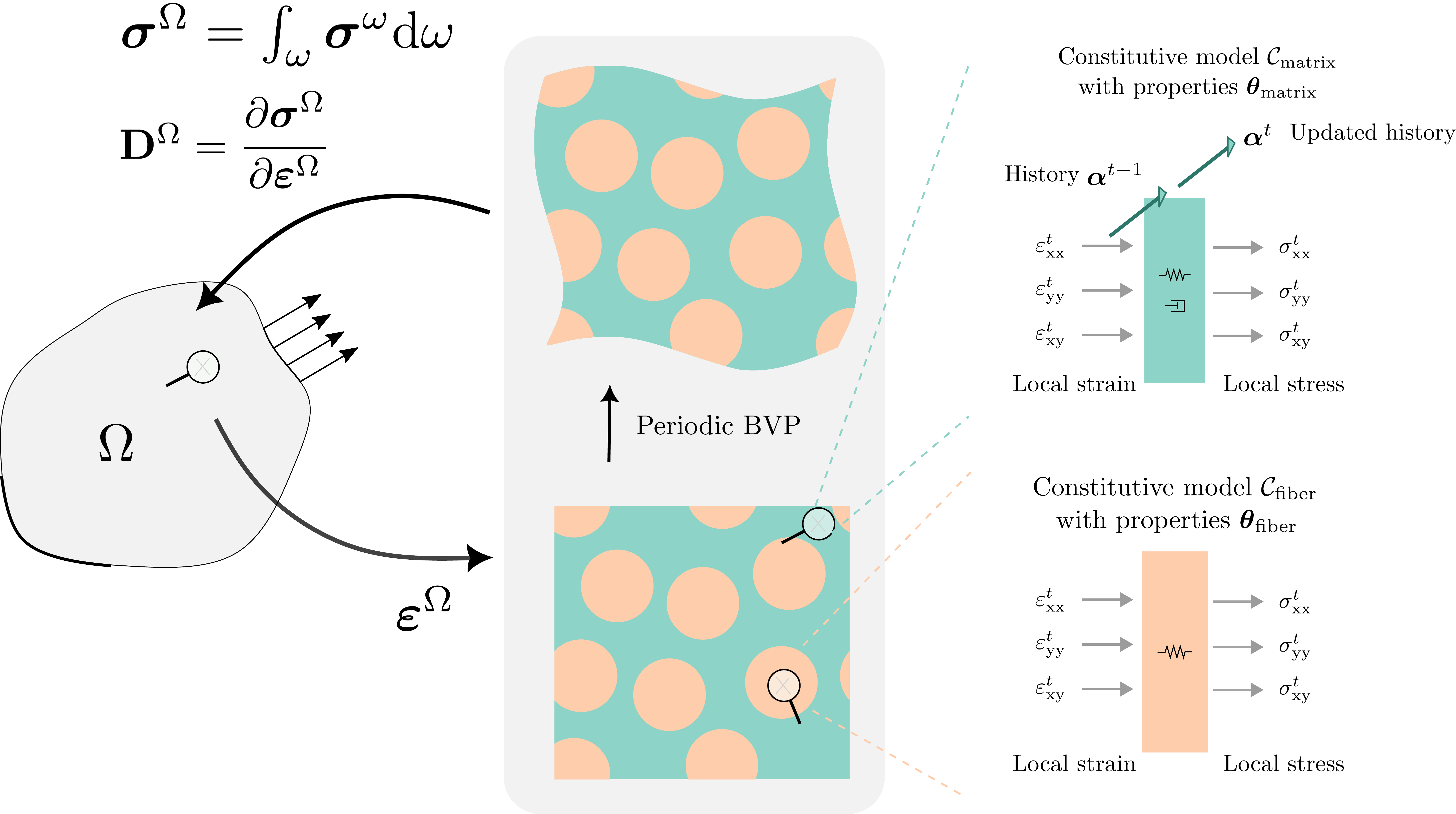}
\caption{FE$^2$ scheme exemplified for composite RVE with elastoplastic matrix and elastic fibers.}
\label{fig:fe2}
\end{figure}

In this formulation, the constitutive behavior of an integration point at the macroscale is not obtained in closed-form, but rather from another nested FE problem at a much smaller scale, here the microscale. This separation between the scales results in a flexible framework suitable for modeling complex material behavior using regular physics-based material models, without any assumptions at the macroscale (\Cref{fig:fe2}, left). The method has three main steps, granted that a suitable Representative Volume Element (RVE) has been identified. The first step consists in downscaling the macroscopic strain $\macrostrainvec$ as appropriate boundary conditions imposed to the RVE. For this, the standard procedure in first-order homogenization relates the displacement field of the micromodel to the macroscopic strain through: 
\begin{equation}
\label{eq:scalebridge}
\mathbf{u}^{\omega} = \boldsymbol{\varepsilon}^{\Omega} \mathbf{x}^{\omega} + \tilde{\mathbf{u}}
\end{equation}
where $\mathbf{x}^{\omega}$ refers to a position inside the RVE and $\tilde{\mathbf{u}}$ is the fluctuation field resulting from the microscopic in-homogeneity. In this work, we consider periodic boundary conditions (PBC) to represent the behavior of the RVE. This type of BC has proven to be versatile and fulfills important requirements in the scale bridging procedure. At the microscale, the constitutive models assigned to each of the phases establish the relation between microscopic strains and stresses, including the update of any internal variables related to time- and/or path-dependent materials (\Cref{fig:fe2}, right). 

Finally, the homogenized stress can be recovered by integrating the stress field over the micromodel domain:
\begin{equation}
\label{eq:macrostress}
\boldsymbol{\sigma}^{\Omega} = \int_{\omega} \boldsymbol{\sigma}^{\omega} \partial \omega \hspace{0.4cm} \Rightarrow \hspace{0.4cm} \boldsymbol{\sigma}^{\Omega} = \cfrac{1}{ \sum_{i=1}^{n} w_i} \, \sum_{i=1}^{n} w_i \, \boldsymbol{\sigma}^{\omega}
\end{equation}
where $w_i$ are the weights from a numerical quadrature, which are strictly positive and sum up to the volume of the micromodel. For an in-depth discussion of the method, we refer to \cite{Geers2017}. Despite its appeal, FE$^2$ is associated with large computational effort, resulting in prohibitive costs even in modern computers. The main bottleneck lies in its nested nature, requiring the solution of multiple BVPs at the microscale for each macroscale solve. In the following sections, we present a brief overview of the constitutive models considered in this work and the surrogate-based multiscale alternative, which has been explored in previous works as an efficient and accurate model. 

\subsection{Constitutive models}
\label{subsec:models}

For the RVE studied in this work, two constitutive models are considered: an elastoplastic model for the  matrix, and a linear elastic model for the fibers, referred to as $\mathcal{C}_{\textrm{matrix}}$ and $\mathcal{C}_{\textrm{fiber}}$, respectively. 

The Helmholtz free energy for the elastoplastic model is given as
\begin{equation}
    \psi = \cfrac{1}{2} \left[ \boldsymbol{\varepsilon}-\boldsymbol{\varepsilon}^{\textrm{p}}\right] : \mathbf{E} : \left[ \boldsymbol{\varepsilon}-\boldsymbol{\varepsilon}^{\textrm{p}}\right] + \cfrac{1}{2} H_\textrm{iso} k^2,
\end{equation}
where $\mathbf{E}$ is the elastic stiffness tensor, $H_\textrm{iso}$ is the isotropic hardening modulus, $\boldsymbol{\varepsilon}$ is the total strain, $\boldsymbol{\varepsilon}^{\textrm{p}}$ is the plastic strain, and $k$ is the isotropic hardening variable. 

The stress, isotropic hardening stress and tangent stiffness matrix are then given as
\begin{equation}
    \boldsymbol{\sigma} = \cfrac{\partial \psi}{\partial\boldsymbol{\varepsilon}} = \mathbf{E} :\left[\boldsymbol{\varepsilon}-\boldsymbol{\varepsilon}^{\textrm{p}}\right], \quad \kappa = -\cfrac{\partial \psi}{\partial k} = - H_\textrm{iso} k, \quad \mathbf{C} = \cfrac{d\boldsymbol{\sigma}}{d\boldsymbol{\varepsilon}} = \mathbf{C}^{\textrm{e}} : \left[ \mathbf{I} - \cfrac{\partial\boldsymbol{\varepsilon}^{\textrm{p}}}{\partial\boldsymbol{\varepsilon}} \right],
\end{equation}
where $\mathbf{I}$ is the identity matrix and the von Mises yield criterion is given by
\begin{equation}
    \Phi = \sqrt{\cfrac{3}{2} \, \boldsymbol{\sigma}^\textrm{dev}:\boldsymbol{\sigma}^\textrm{dev}} - [\sigma_{\textrm{h}_0} + \kappa], \quad \boldsymbol{\sigma}^\textrm{dev} := \boldsymbol{\sigma} - \cfrac{1}{3} \, \textrm{tr}(\boldsymbol{\sigma}) \, \mathbf{I}.
\end{equation}
The associative evolution of plastic strains is assumed, and a nonlinear, saturating evolution of the isotropic hardening, resulting in the Voce hardening law, are chosen
\begin{equation}
\dot{\boldsymbol{\varepsilon}}^\textrm{p} = \dot{\lambda} \cfrac{\partial \Phi}{\partial \boldsymbol{\sigma}}, \quad
\dot{k} = -\dot{\lambda}\left[1 - \frac{\kappa}{\kappa_\infty}\right],
\end{equation}
where $\dot{\lambda}$ is the increment in plastic multiplier. Finally, for rate independent plasticity, the KKT-conditions must be fulfilled
\begin{equation}
    \Phi \leq 0, \quad \dot{\lambda} \geq 0, \quad \Phi\ \dot{\lambda} = 0.
\end{equation}
We remark that this results in the exponentially saturating isotropic hardening
\begin{align}
    \sigma_\textrm{h} = \sigma_{\textrm{h}_0} + \kappa = 
    \sigma_{\textrm{h}_0} + \kappa_\infty \left[1 - \exp\left(\frac{H_\textrm{iso}}{\kappa_\infty}\lambda\right)\right].
\end{align}

For the elastic model, the energy, stress and tangent stiffness matrix simplify to:
\begin{equation}
\label{eq:elasticforms}
\psi = \cfrac{1}{2} \, \boldsymbol{\varepsilon} : \mathbf{E} : \boldsymbol{\varepsilon}, \quad
\boldsymbol{\sigma}  = \cfrac{d \psi}{d \boldsymbol{\varepsilon}} = \mathbf{E} : \boldsymbol{\varepsilon}, \quad \mathbf{C} = \cfrac{d{\boldsymbol{\sigma}}}{d\boldsymbol{\varepsilon}} =  \mathbf{C}^{\textrm{e}}.
\end{equation}
Finally, it is worth highlighting that $\mathbf{C}^{\textrm{e}}$ and $\mathbf{C}$ are Symmetric Positive Definite (SPD) matrices for both $\mathcal{C}_{\textrm{matrix}}$ and $\mathcal{C}_{\textrm{fiber}}$. For elasticity this is trivially satisfied, while for Von Mises plasticity it is guaranteed through the choice for an associative plastic flow and a (positive and convex) hardening function.

\subsection{Replacing the micromodel with an approximate model}
\label{subsec:prnn}

We now substitute the original microscopic BVP model of \cref{subsec:micro} with a Physically Recurrent Neural Network (PRNN). In this section, we briefly discuss the main elements of PRNNs and highlight the features explored in this work. For details on implementation and training aspects, we refer to \cite{Maiaetal2023}. The fundamental difference between this hybrid network and traditional data-driven NNs and/or other highly-tailored architectures that ensure, \textit{e.g.}, thermodynamic consistency by construction, is the use of intact constitutive models instead of regular activation functions and memory mechanisms, such as gates in RNNs and self-attention in transformers. With this, the complex time- and path-dependent constitutive behaviors are preserved and reproduced automatically, leaving the remaining parts of the network the task to best combine their response. 
\begin{figure}[!b]
\centering
 \includegraphics[width=.7\textwidth]{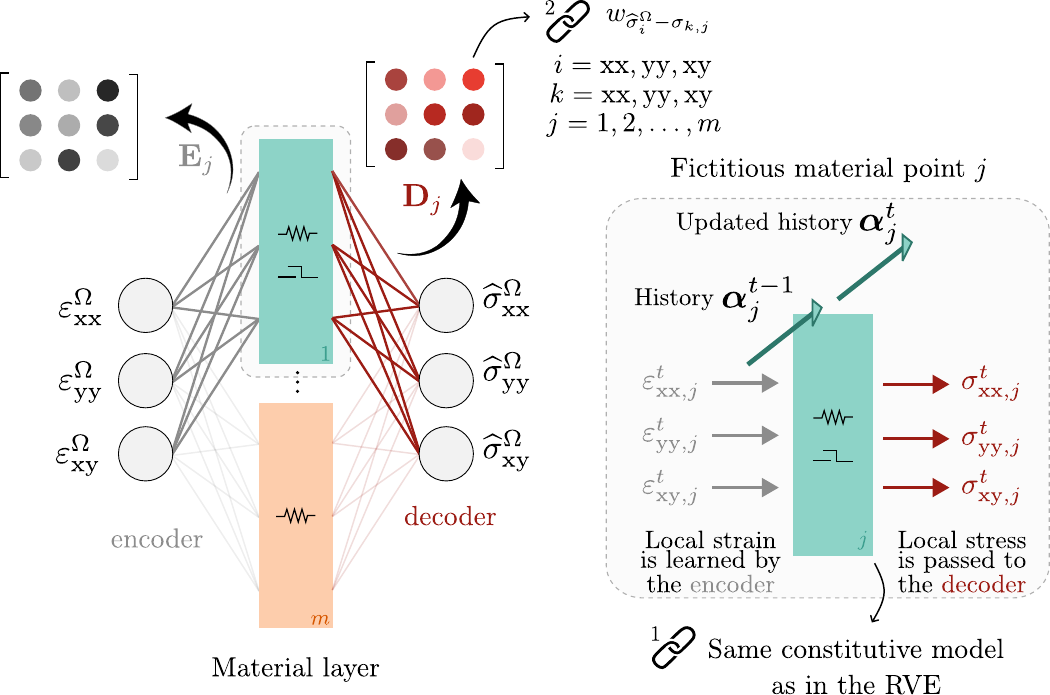}
\caption{Classical PRNN architecture with dense connections and elastoplastic models embedded in the material layer.}
\label{fig:prnnbasic}
\end{figure}

In the PRNN, the constitutive models are embedded in what we call a \textit{material layer}, which stands between an \textit{encoder} and a \textit{decoder}, as illustrated in \Cref{fig:prnnbasic}. First, the values obtained by the \textit{encoder} are grouped in sets of the same size as the macroscopic strains, referred as \textit{fictitious material points}, and interpreted as local strains, giving rise to the first link with \Cref{fig:fe2}. In this design, each fictitious material point is assigned to one of the constitutive models adopted in the full-order solution. This way, the constitutive model directly computes the stress and the internal variables (if any), removing from the data-driven parts of the network the responsibility to learn complex time and path-dependent constitutive laws. For illustration purposes, we show a network with the same two models as in \Cref{fig:fe2}. In an analogy with the homogenization procedure, the encoder corresponds to the solution of the RVE boundary value problem formulated in terms of the macroscopic strain $\macrostrainvec$ \cite{Geers2017}, while the evaluations at the material layer correspond to the recovery of the full-field of stresses. 

The final ingredient of PRNNs is the \textit{decoder}. This component is responsible for transforming the set of local stresses obtained from the fictitious material points into the homogenized stress $\macrostressvec$. Drawing inspiration from the numerical integration procedure in \Cref{eq:macrostress}, we establish the second link with \Cref{fig:fe2}, in which we impose a constraint on the positivity of the weights of the decoder by taking the absolute value of the unconstrained weights during the optimization and evaluation. For training the network, the Mean Squared Error (MSE) loss function is adopted:
\begin{equation}
\label{eq:lossavg}
\ell_{\textrm{avg.}} = \frac{1}{N} \ \sum_{i=1}^{N} \frac{1}{2} \ \| \ \macrostressvec(\boldsymbol{\varepsilon}^{\Omega}_i) - \widehat{\boldsymbol{\sigma}}^{\Omega}(\boldsymbol{\varepsilon}^{\Omega}_i) \ \|^2  
\end{equation}
where $N$ is the number of snapshots. 

It is worth mentioning that both the encoder and decoder can take the form of an arbitrary multilayer perceptron, as shown in \cite{Maiaetal2023}. Here we narrow the architecture of these two components to a simple linear relation and focus on model variations tailored for low-data situations that can lead to better interpretability and/or guaranteed thermodynamic consistency. Details on these architecture variations are discussed in the following. 

\section{Architectural choices for interpretability and thermodynamical consistency}
\label{sec:decoder}
This section explores different variations of the decoder architecture, starting from the baseline dense design in \Cref{fig:prnnbasic}. The proposed changes are motivated in terms of interpretability, flexibility and thermodynamic consistency. Throughout this work, we further discuss how these architectural choices affect the learned latent space and the resulting homogenized response.

\subsection{Towards sparsity}
\label{subsec:sparsity}

\begin{figure}[!ht]
\centering
\includegraphics[width=.87\textwidth]{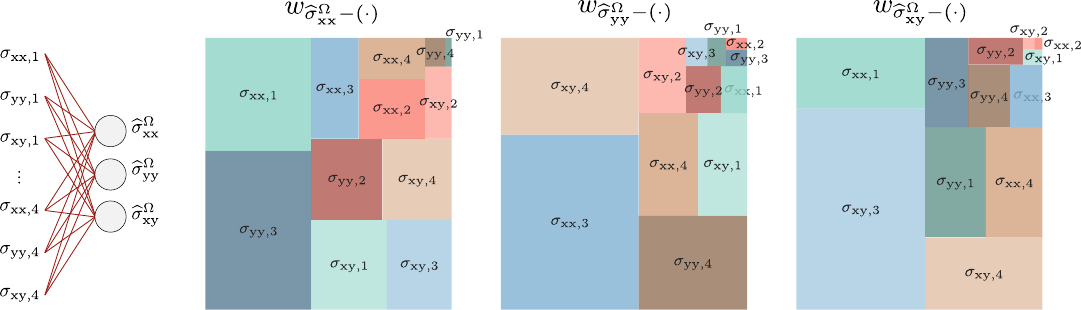}
\caption{Representation of weights of PRNN with 4 material points: full connectivity between output and material layer allows high expressivity\label{fig:decdense}.}
\label{fig:decodersp5}
\end{figure}

Firstly, we visualize the decoder weights of a PRNN with dense connections through the lens of squarified treemaps. Details on model selection are omitted for the present discussion but can be found in \Cref{sec:accuracylowdata}, where all alternatives are assessed in terms of accuracy. In the treemaps, each rectangle represents a weight used to compute a given component of the homogenized stress. Specifically, the area of each rectangle is proportional to the magnitude of the corresponding weight, which is not to be confused with the magnitude of the stress computed by the constitutive model itself. In \Cref{fig:decdense}, we use a network with four material points, all with $\mathcal{C}_{\textrm{matrix}}$ assigned as their constitutive models, for illustration purposes. Due to the full connectivity between the output and material layer, all three stress components from the four material points are used when computing $\widehat{\sigma}_i^{\Omega}$, resulting in a treemap with 12 rectangles. 

Note that for each treemap, different weights are assigned to the same material point output depending on the macroscopic stress component being approximated. In this particular example, the largest weight contributing to $\predmacrostress_{\textrm{xx}}$ originates from the $y$-component of fictitious point 3, while for $\predmacrostress_{\textrm{yy}}$ it originates from the $x$-component. Although this introduces flexibility to the surrogate model, this behavior is inconsistent with the true homogenization process. While the size of the rectangles related to cross-components do not necessarily imply they have the highest contribution in the approximate homogenized stress, since that also depends on the stress magnitude from the fictitious points, a single integration weight is assigned per material point in the full-order homogenization and cross-component stresses do not contribute at all. These observations motivate the modifications proposed in this work. 
\begin{figure}[!b]
\centering
 \includegraphics[width=.6\textwidth]{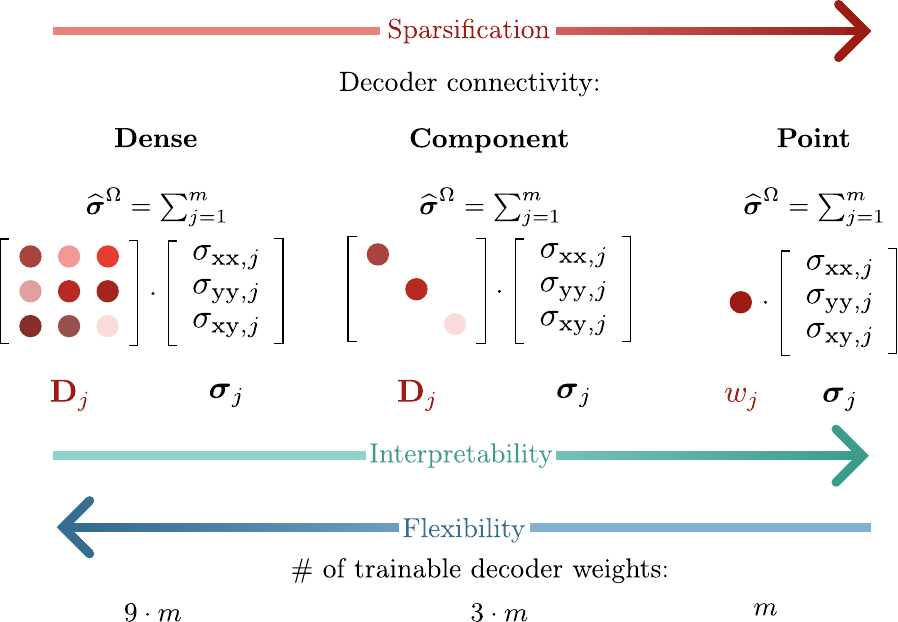}
\caption{Decoder connectivities investigated in this work and their degree of sparsification, interpretability and flexibility, with $m$ being the number of fictitious material points.}
\label{fig:prnndecoders}
\end{figure}

The first modification we study is the removal of the cross-component connections, allowing only matching components to contribute to the predicted homogenized stress. Next, we impose that all stress components from a material point share the same weight. The latter is the closest to the homogenization procedure described in \Cref{eq:macrostress}, where integration weights are assigned per material point, not per stress component. This is the least flexible architecture but the most interpretable one. These three options, including the baseline dense decoder are hereafter referred to as \Densedec, \Componentdec \ and \Pointdec, and are illustrated in \Cref{fig:prnndecoders}, where the number of trainable parameters of the decoder is included for reference. In all settings, the absolute function is applied to all the trainable parameters of the decoder to ensure positivity.  

In addition to these changes in the connectivity, we investigate the use of a normalization condition in which the weights that connect a given component of the homogenized stress tensor to the contributing stresses from the fictitious points are constrained to sum to one. This condition allows weights to be interpreted as relative contributions, establishing a direct analogy with the volume averaging in \Cref{eq:macrostress}, where weights used for the numerical integration of the stresses are strictly positive with their sum totalling the volume of the RVE. Furthermore, unlike common regularization strategies based on penalty parameters, such as L1 or L2, this approach needs no definition of additional hyper-parameter. 

\subsection{Thermodynamic consistency}
\label{subsec:thermo}

\begin{figure}[!b]
\centering
 \includegraphics[width=.65\textwidth]{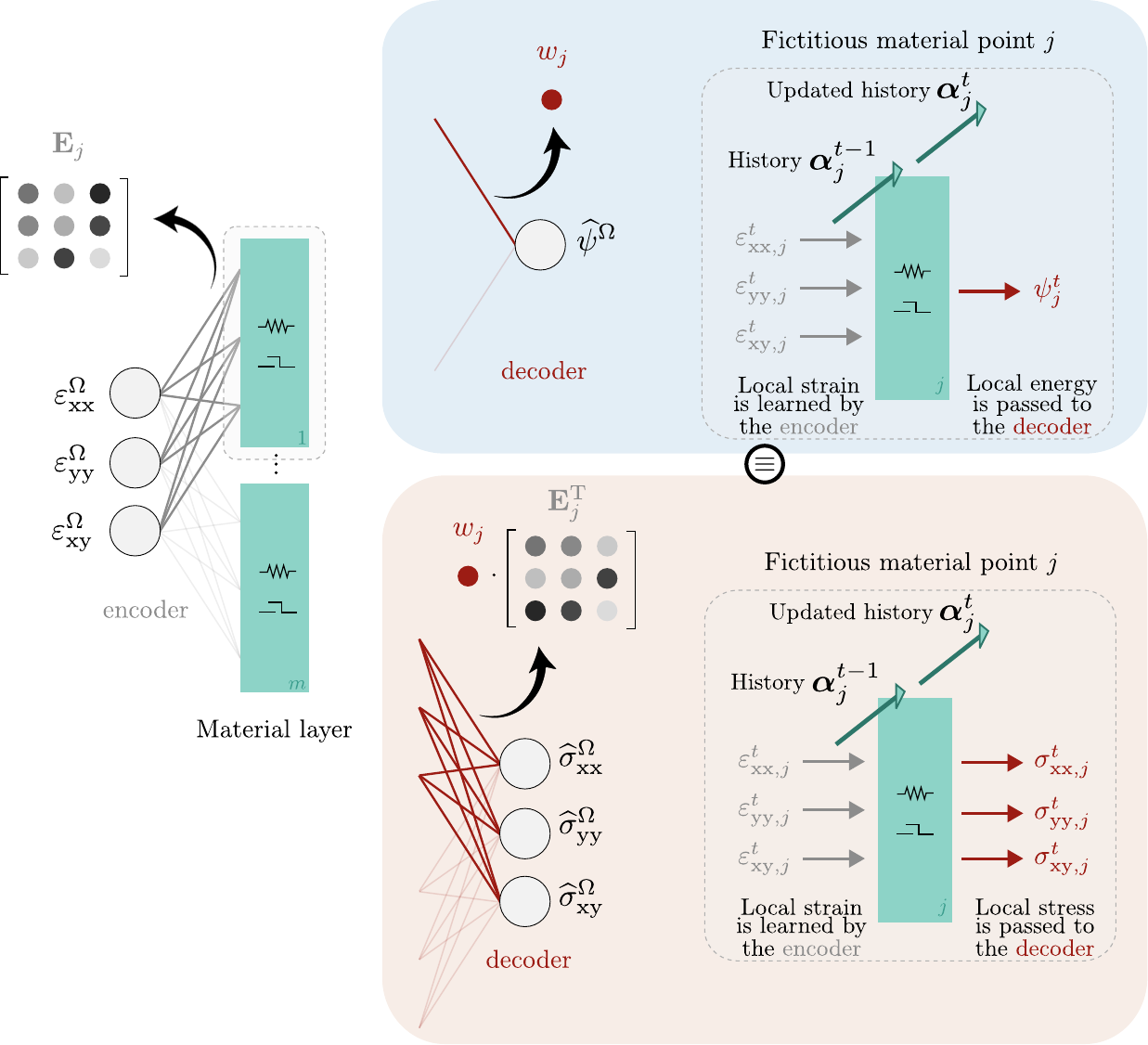}
\caption{Thermodynamically consistent and equivalent PRNN architectures: light-blue box implements an energy-based network, while light-brown box corresponds to the conventional stress-based formulation.}
\label{fig:tcprnn}
\end{figure}

Thus far, we have explored aspects such as flexibility and interpretability. Now, we shift our focus to another key aspect for  robustness in FE$^2$ applications, the thermodynamic consistency between scales. One way to naturally enforce it is to directly derive our architecture from an energy-based formulation. In this case, we would learn to homogenize the local energies from fictitious material points instead of local stresses. This new setting is depicted in the upper (light-blue) box in \Cref{fig:tcprnn}. Based on that design, we can write the macroscopic Helmholtz's free energy as the sum of the energy of all fictitious material points
\begin{equation}
    \label{eq:tcprnnenergy}
    \widehat{\psi}^{\Omega} \left(\macrostrainvec \right) =  \sum_{j=1}^{m} \, w_j \, \psi_j  \left(  \boldsymbol{\varepsilon}_j , \intvarvec_j \right), \quad \boldsymbol{\varepsilon}_j := \mathbf{E}_j \cdot 
    \macrostrainvec \\
\end{equation}
where $\psi_j$ is the free energy of the fictitious material point $j$, taken here as a general history-dependent model with internal variables $\boldsymbol{\alpha}$, and $\mathbf{E}_j$ are the encoder weights associated with the point. From this definition, the macroscopic stress can be derived as
\begin{equation}
\label{eq:stressfromenergy}
\begin{aligned}
\widehat{\boldsymbol{\sigma}}^{\Omega} & = \cfrac{\partial\widehat{\psi}^{\Omega}}{\partial \macrostrainvec} = \sum_{j=1}^{m} w_j \, \cfrac{\partial \psi_{j}}{\partial \boldsymbol{\varepsilon}_j} \, \cfrac{\partial\boldsymbol{\varepsilon}_j}{\partial \macrostrainvec}.
\end{aligned}
\end{equation}
Using that $\boldsymbol{\sigma}_j = \partial \psi_{j}/\partial \boldsymbol{\varepsilon}_j$ and $\mathbf{E}_j = \partial\boldsymbol{\varepsilon}_j/\partial \macrostrainvec$
allow us to transition to a stress-based formulation equivalent to the energy-based with which we started:
\begin{equation}
\label{eq:stressenergyfinal}
\widehat{\boldsymbol{\sigma}}^{\Omega} = \sum_{j=1}^{m} w_j \, \mathbf{E}_j^{\textrm{T}} \cdot \boldsymbol{\sigma}_j.
\end{equation}
This new design has the same number of trainable parameters as the \Pointdec \ discussed previously since the decoder weights are merely scaled versions of the transpose of the encoder. We continue using the normalization proposed in \Cref{subsec:sparsity} to impose the sum-to-one constraint on the trainable decoder weights, and refer to networks with this setting as \NormTCdec \ (from Normalized Thermodynamically Consistent).

With this adjoint encoder-decoder structure, we further expand the applicability of this approach to deal with constitutive models formulated in terms of energy. The two formulations introduced in this section are illustrated in \Cref{fig:tcprnn}. For the numerical examples, we train our networks on homogenized strain-stress pairs as in \Cref{eq:stressenergyfinal}. In the following, we explicitly demonstrate that the Hill-Mandel principle and the positiveness of the dissipation are fulfilled by design, and contrast it with the other alternatives, for which these properties are not guaranteed. 

The Hill-Mandel principle states that the variation of macroscopic energy should equal the volume average of microscopic energy. In the full-order micromodel, this condition reads as
\begin{equation}
\label{eq:hillmandelhf}
	\macrostressvec : \dotmacrostrainvec = \int_{\omega} \microstressvec :\dot{\microstrainvec} \, \textrm{d}\omega,
\end{equation}
while the network counterpart reads as
\begin{equation}
\label{eq:hillmandelnn}
	\predmacrostress  :  \dotmacrostrainvec = \sum_{j=1}^{m} w_j \, \boldsymbol{\sigma}_{j}  :  \dot{\boldsymbol{\varepsilon}}_{j}.
\end{equation}
Expanding the expressions of the predicted macroscopic stress on the left hand-side of \Cref{eq:hillmandelnn} and the expression of the localized fictitious strain on the right-hand side with their dependence on the encoder and decoder parameters, we arrive at
\begin{equation}
	\label{eq:hillmandelpred}
	\left(\sum_{j=1}^{m} w_j \, \mathbf{D}_j \cdot \boldsymbol{\sigma}_j \,\right) \, : \, \dotmacrostrainvec  = \sum_{j=1}^{m} w_j \, \boldsymbol{\sigma}_j \, : \, \left( \mathbf{E}_j \cdot \dotmacrostrainvec \right).
\end{equation}
Rearranging the terms on the left side of \Cref{eq:hillmandelpred}, we obtain
\begin{equation}
\label{eq:hillmandelstep}
	\sum_{j=1}^{m} w_j \, \boldsymbol{\sigma}_j \, : \, \left( \mathbf{D}_j^{\textrm{T}} : \dot{\boldsymbol{\varepsilon}}^{\Omega}  \right) = \sum_{j=1}^{m} w_j \, \boldsymbol{\sigma}_j \, : \, \left( \mathbf{E}_j : \dotmacrostrainvec \right).
\end{equation}
To ensure this condition is fulfilled, we need to enforce $\mathbf{E}_j = \mathbf{D}_j^{\textrm{T}}$, which results in the adjoint encoder-decoder structure naturally derived from the energy formulation discussed previously. In this case, $\mathbf{E}_j$ does not need to follow any particular form, but $w_j$ needs to be positive. As a consequence, architectures that do not follow this adjoint structure, including the variations presented in \Cref{subsec:sparsity}, do not fulfil the Hill-Mandel condition, and therefore do not guarantee energetic consistency despite producing accurate stress predictions.

Based on these definitions, it follows that the macroscopic tangent stiffness matrix given by the network is
\begin{equation}
\label{eq:eq:macrotangentconsistent}
	\widehat{\mathbf{C}}^{\Omega} = \sum_{j=1}^{m} \, w_j \, \mathbf{E}^{\textrm{T}}_j : \mathbf{C}_j : \mathbf{E}_j,
\end{equation}
where $\mathbf{C}_j$ is the tangent from material point $j$. In the examples explored in this contribution, the true macroscopic tangent stiffness matrix $\mathbf{C}^{\Omega}$ is SPD, as a result of the chosen constitutive models. Unlike the previous architecture, this adjoint arrangement ensures that the network counterpart $\widehat{\mathbf{C}}^{\Omega}$ will also inherit that property if $\mathbf{C}_j$ is SPD and $w_j>0$.

The second aspect to be discussed is the fulfilling the dissipation inequality,
\begin{align}
    \label{eq:diss}
    \mathcal{D} = \boldsymbol{\sigma}:\dot{\boldsymbol{\varepsilon}} - \dot{\psi} \geq 0
\end{align}
The dissipation for the network model is,
\begin{align}
\widehat{\mathcal{D}}^{\Omega} &= 
\widehat{\boldsymbol{\sigma}}^{\Omega} : \dot{\boldsymbol{\varepsilon}}^{\Omega} - 
\dot{\widehat{\psi}}{}^{\Omega}
\end{align}
Inserting the stress from Equation \eqref{eq:stressenergyfinal} and the free energy from Equation \eqref{eq:tcprnnenergy} gives
\begin{equation}
\label{eq:dissprnn}
\widehat{\mathcal{D}}^{\Omega} = \sum_{j=1}^{m} w_j \, \boldsymbol{\sigma}_j : \dot{\boldsymbol{\varepsilon}}_j - \sum_{j=1}^{m} \, w_j \, \dot{\psi}_j = \sum_{j=1}^{m} w_j \mathcal{D}_j
\end{equation}
where $\mathcal{D}_j = \boldsymbol{\sigma}_j : \dot{\boldsymbol{\varepsilon}}_j - \dot{\psi}_j$ is the dissipation for the material model at material point $j$. As $w_j \geq 0$, the network model is thermodynamically consistent for thermodynamically consistent material models at each material point.

Note that while the dissipation is well-defined at the material level regardless of connectivity and normalization, the macroscopic quantity is not. For architectures such as \Densedec \ and \Componentdec, there is no natural way to aggregate a scalar-valued quantity with weights that vary per component. In contrast, the \Pointdec \ and \TCdec \ (and their normalized versions) admit a straight-forward and consistent weighted sum over the points, as in \Cref{eq:dissprnn}. However, only the \TCdec \ satisfies both the energetic consistency and the non-negative dissipation at the same time. 

\section{Accuracy under severe data scarcity}
\label{sec:accuracylowdata}

The new PRNN architectures we propose have a significantly higher degree of inductive bias than the original dense architecture in \cite{Maiaetal2023}. It is therefore interesting to investigate their behavior under extremely low data scenarios comprising between one and five short stress-strain training paths. We consider different material layer sizes ranging from 2 to 30 fictitious material points, and for each combination of model complexity and training set size, 10 initializations are considered. In each initialization, a small training set is randomly drawn from a pool of \num{1000} paths. The data generation procedure follows the approach detailed in \cite{Maiaetal2023}, including material properties and RVE geometry, with GP-based paths being the preferred choice due to their higher complexity. Fixed validation $\mathcal{V}_{\textrm{GP}}$ and test $\testunlgp$ sets consisting of 150 GP-based paths each are considered for the hyper-parameter study and for assessing the network in terms of accuracy. 

The Adam algorithm is used to optimize the network's parameters with two stopping criteria. The first is a maximum number of epochs, set to \num{8000}, and the second is an early stopping criterion which interrupts the training if there is no improvement in the validation loss for 400 consecutive epochs. The batch size is set to 1 path, while the remaining optimization parameters are  fixed to Pytorch defaults. Training and validation losses are calculated using the Mean Squared Error (MSE), while model performance on the test set is assessed using the Mean Absolute Error (MAE) and its relative counterpart (RMAE) for easier interpretability.

Finally, following the idea that the decoder weights represent a homogenization operator, we modify the initialization of this set of parameters in all of the alternatives, except for \TCdec \ and \ \NormTCdec, to:
\begin{enumerate}[label=\arabic*)]
\item draw weights from an uniform distribution between 0 and 1;
\item normalizing weights connecting to $\widehat{\sigma}^{\Omega}_{i}$, $w_{{\widehat{\sigma}^{\Omega}_{i}}-(\cdot)}$, to sum to 1.
\end{enumerate}
Note that this procedure normalizes the decoder weights only at the \textit{start} of the training procedure. For unnormalized decoders, these weights are free to change and to sum to arbitrary (positive) values as epochs evolve, while for normalized decoders, the sum-to-one instruction is a constraint fulfilled \textit{at all times}. In the following, we start with the unnormalized and then move to the normalized decoders, including the thermodynamically consistent architecture, to disentangle potential accuracy gains stemming from two distinct aspects: sparsity and weight normalization.

\subsection{Unnormalized decoders}
\label{subsec:unnormaccuracy}

\begin{figure}[!h]
\centering
\includegraphics[width=.9\textwidth]{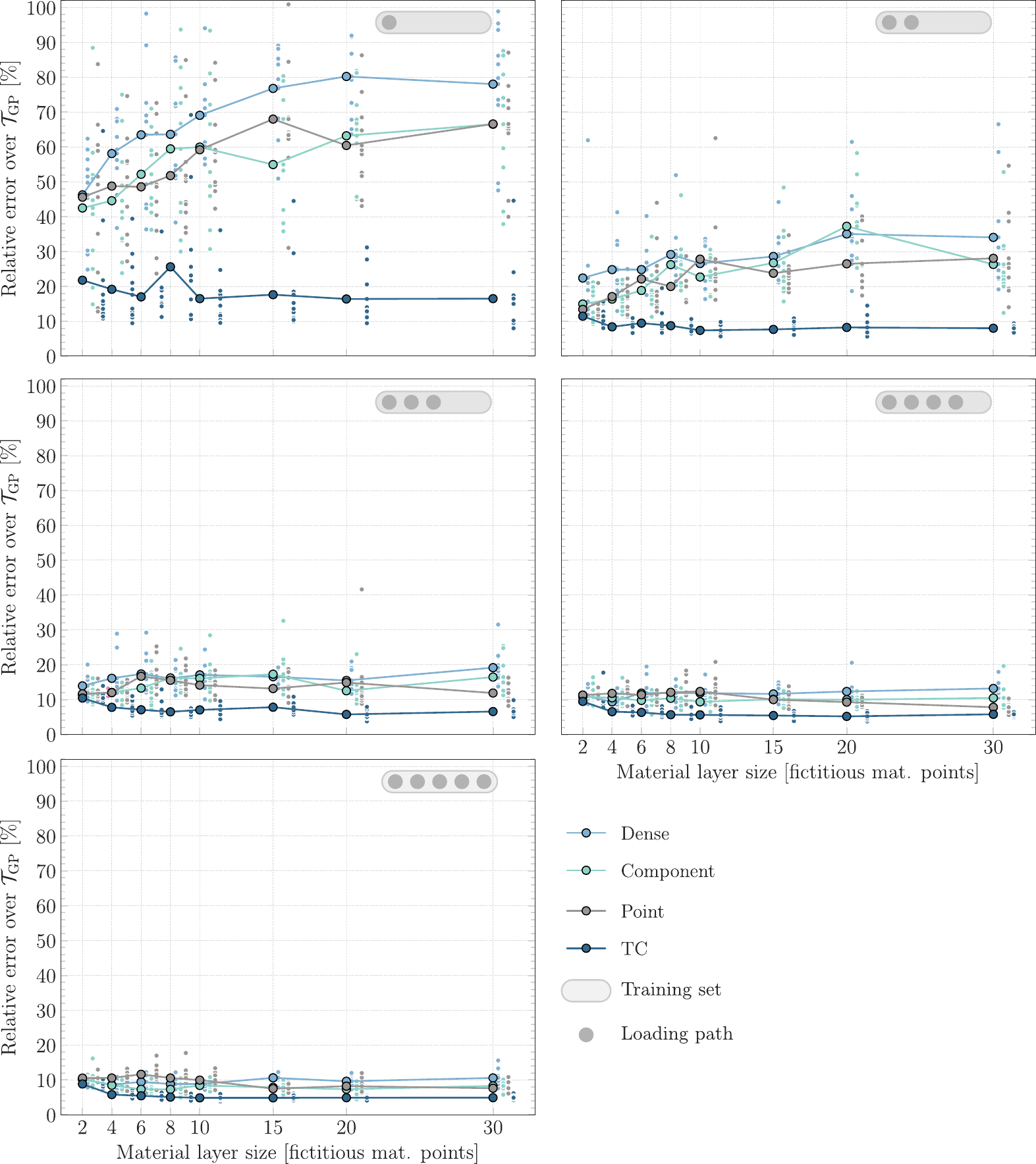} 
\caption{Relative error over $\testunlgp$ for different material layer sizes  and training set sizes considering unnormalized decoders..}
\label{fig:relerrortestnonnorm}
\end{figure}

In \Cref{fig:relerrortestnonnorm}, we plot the average relative errors of the trained networks over the test set $\testunlgp$ for all combinations of material layer and training set sizes. Each color corresponds to a different decoder architecture and each solid line corresponds to the average error of the different initializations (dots spread out vertically). One observation from this plot concerns the marked difficulty to generalize from limited data, illustrated by the steep growth of errors as the number of training curves decreases. 

As the decoder connectivity is reduced, the network incorporates stronger inductive biases, more closely mirroring the RVE homogenization. In \Cref{fig:relerrortestnonnorm}, the lines associated to the sparse decoders are virtually always below the one associated to the \Densedec \ one. With smaller networks (2-6 fictitious mat. points), the addition of more curves to the training set result in a significant accuracy gain and variance reduction, consistently reaching the 10\% mark with 4 training curves. With 5 curves, the accuracy of the four decoders converge towards the same level. Overall, sparse architectures improve performance in the low-data regime, although training on 1-2 curves remains challenging.
\begin{figure}[!h]
\centering
\begin{minipage}{.45\textwidth}
\subfloat[\Densedec, 15 fictitious mat. points (270 weights)\label{fig:ls15tr01dense}]{\includegraphics[width=\textwidth]{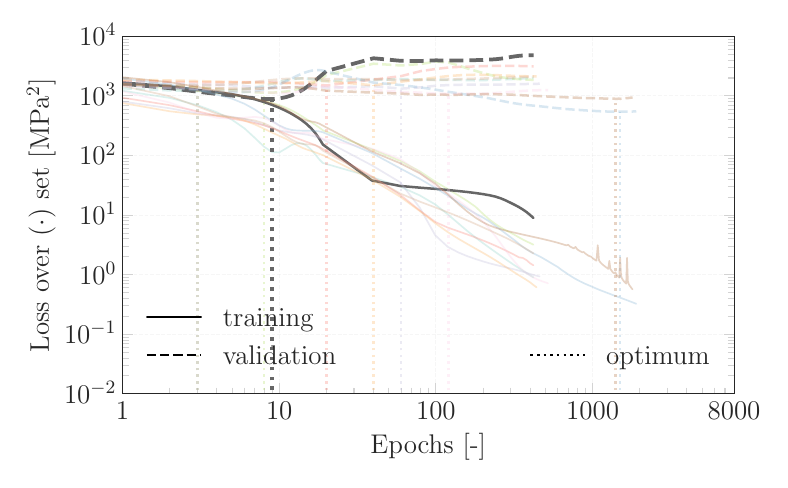}}
\end{minipage}
\begin{minipage}{.45\textwidth}
\subfloat[\TCdec, 30 fictitious mat. points (300 weights)\label{fig:ls30tr01tc}]{\includegraphics[width=\textwidth]{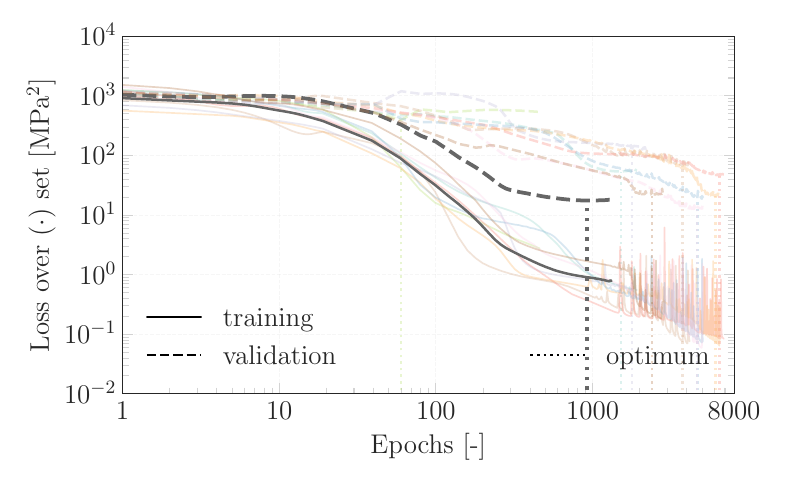}}
\end{minipage}
\caption{Training and validation losses from different initializations of PRNNs with \Densedec \ and \TCdec \ decoders trained on one curve.}
\label{fig:losses}
\end{figure}

\begin{figure}[!b]
\centering
\begin{minipage}{.8\textwidth}
\centering
\subfloat[Strain view of paths used to train (black) and test (cobalt)\label{fig:spherels30tr01net07}]{\includegraphics[width=\textwidth]{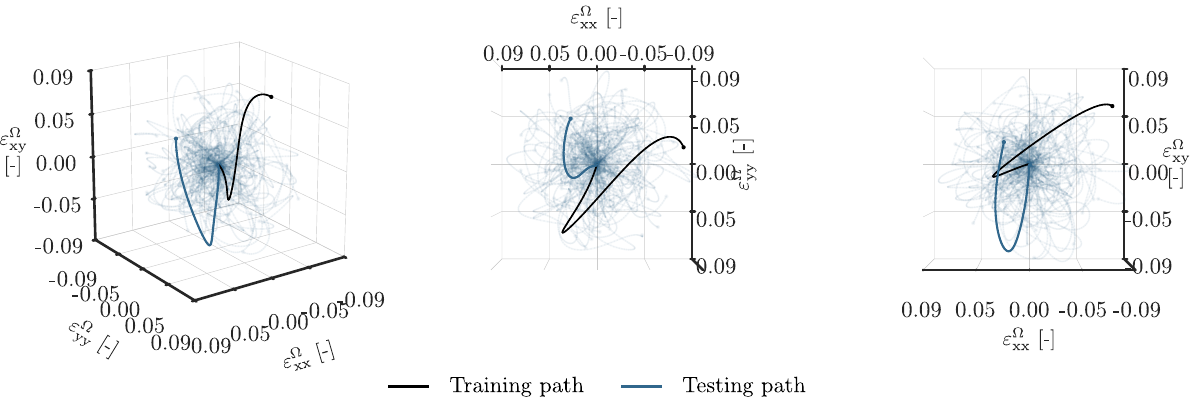}}
\end{minipage}
\begin{minipage}{.8\textwidth}
\centering
\subfloat[Stress-strain curve of testing path highlighted in \Cref{fig:spherels30tr01net07}\label{fig:stressstrainls30tr01net07}]{\includegraphics[width=\textwidth]{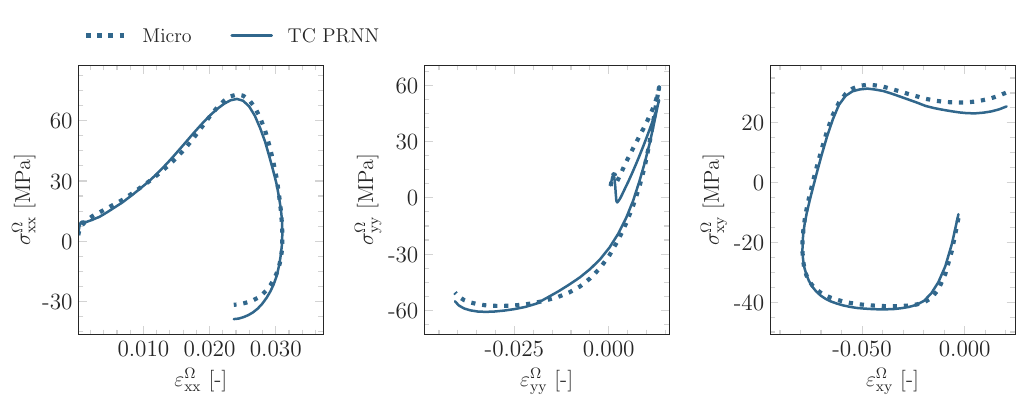}}
\end{minipage}
\caption{Training and test paths for PRNN with \TCdec \ decoder and 30 material points trained on one loading path.}
\label{fig:errortest}
\end{figure}

The \TCdec \ architecture is the exception in this case. \TCdec \ networks trained on a single curve provide accurate predictions comparable to those obtained with larger sets using any of the other alternatives. To further illustrate this difference, we compare in \Cref{fig:losses} the training and validation losses for \Densedec \ and \TCdec \ networks with similar number of trainable parameters. Note that, given enough epochs, the \Densedec \ networks eventually fine-tune their predictions over the training curve, which often compromises their generalization power. Even though we employ an early stopping strategy, the amount of (training) data is still insufficient for those networks to reliably extract any pattern, resulting in consistently high validation errors. 

In contrast, \TCdec \ networks improve training and validation together over the epochs, with no signs of overfitting even under severe data scarsity. To illustrate the accuracy of these models in predicting the homogenized stresses, we consider the one with the lowest validation error. The training and testing paths are shown in \Cref{fig:spherels30tr01net07}. Note that despite covering a limited and distinct region of the design space, the model trained on a single curve already produces smooth homogenized stresses over the entire loading space, as illustrated by the representative path in \Cref{fig:stressstrainls30tr01net07}.
\begin{figure}[!b]
\centering
\begin{minipage}{.445\textwidth}
\subfloat[Homogenized stresses and plastic strains\label{fig:dissmacro}]{
\includegraphics[valign = b, width=\textwidth]{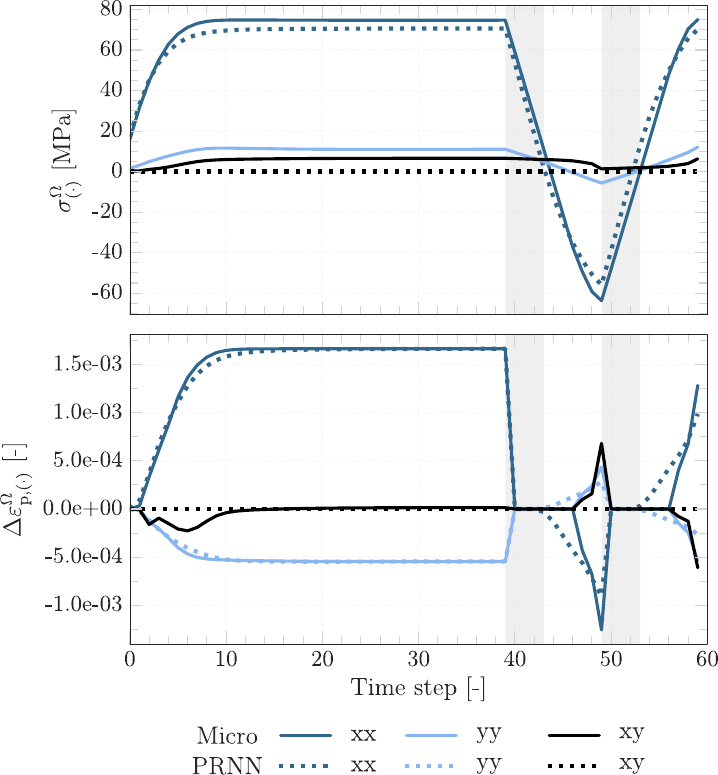}\vphantom{\includegraphics[width=0.57\textwidth,valign=b]{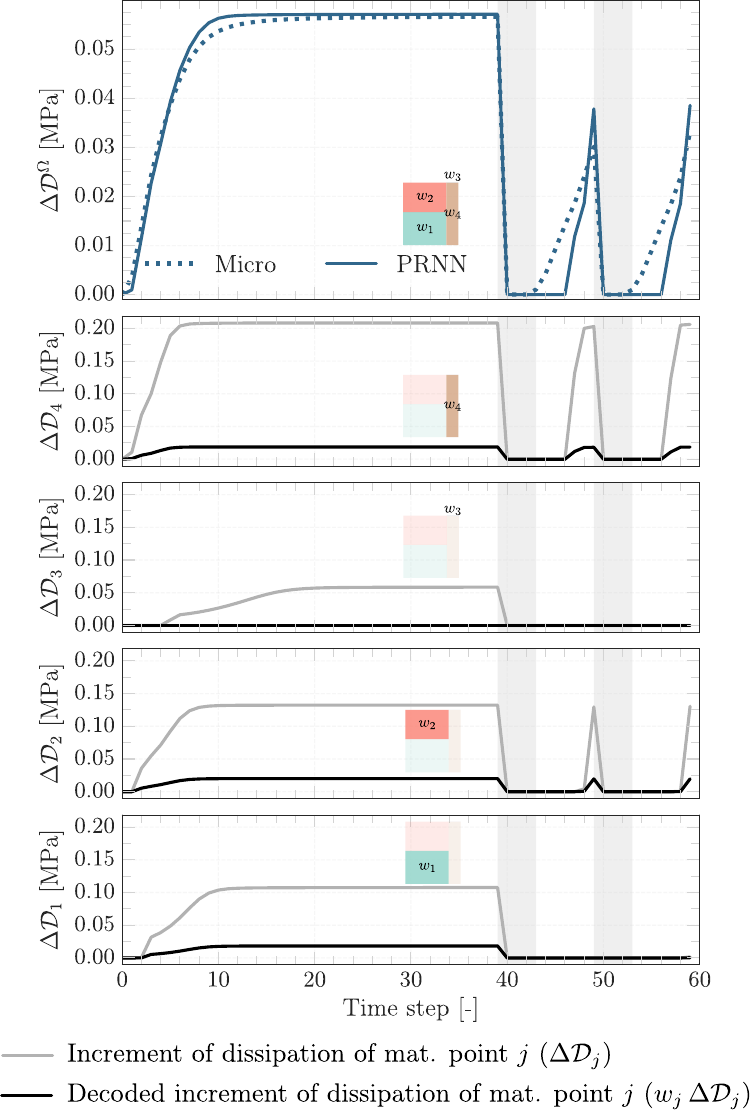}}}
\end{minipage}
\hfill
\begin{minipage}{.545\textwidth}
\subfloat[Homogenized and local increments of dissipation\label{fig:dissprnn}]{
\includegraphics[valign = b, width=\textwidth]{diss_localpts.pdf}}
\end{minipage}
\caption{Macroscopic increment of dissipation from representative proportional loading path with two unloading phases (light gray zones) computed with micromodel and PRNN with \TCdec \ trained on one GP-based path and 4 material points.}
\label{fig:disscurve}
\end{figure} 

Finally, to illustrate the thermodynamical consistency of the \TCdec \ networks, we consider the evolution of the increment of dissipation. For easier visualization, instead of GP-based paths, this time we examine a proportional loading path with two unloading phases. In \Cref{fig:dissmacro}, we plot the macroscopic increments of dissipation and plastic strains, highlighting the unloading phases in light gray. As expected, while unloading takes place, no plastic strains develop in the micromodel or the PRNN. In \Cref{fig:dissprnn}, we plot the macroscopic increment of dissipation and the corresponding decoded quantities per fictitious material point. Since the increment from each point is always $\ge 0$, as well as the decoder weight associated with it, the macroscopic increment of dissipation is non-negative by design. It is also worth noting that, similarly to the micromodel, the fictitious points do not all contribute to dissipation at the same time. For example, during reloading from time step 43 to 49, points 2 and 4 develop plasticity, while points 1 and 3 remain elastic.

\subsection{Normalized decoders}
\label{subsec:weightnorm}
\begin{figure}[!b]
\centering
\includegraphics[width=.8\textwidth]{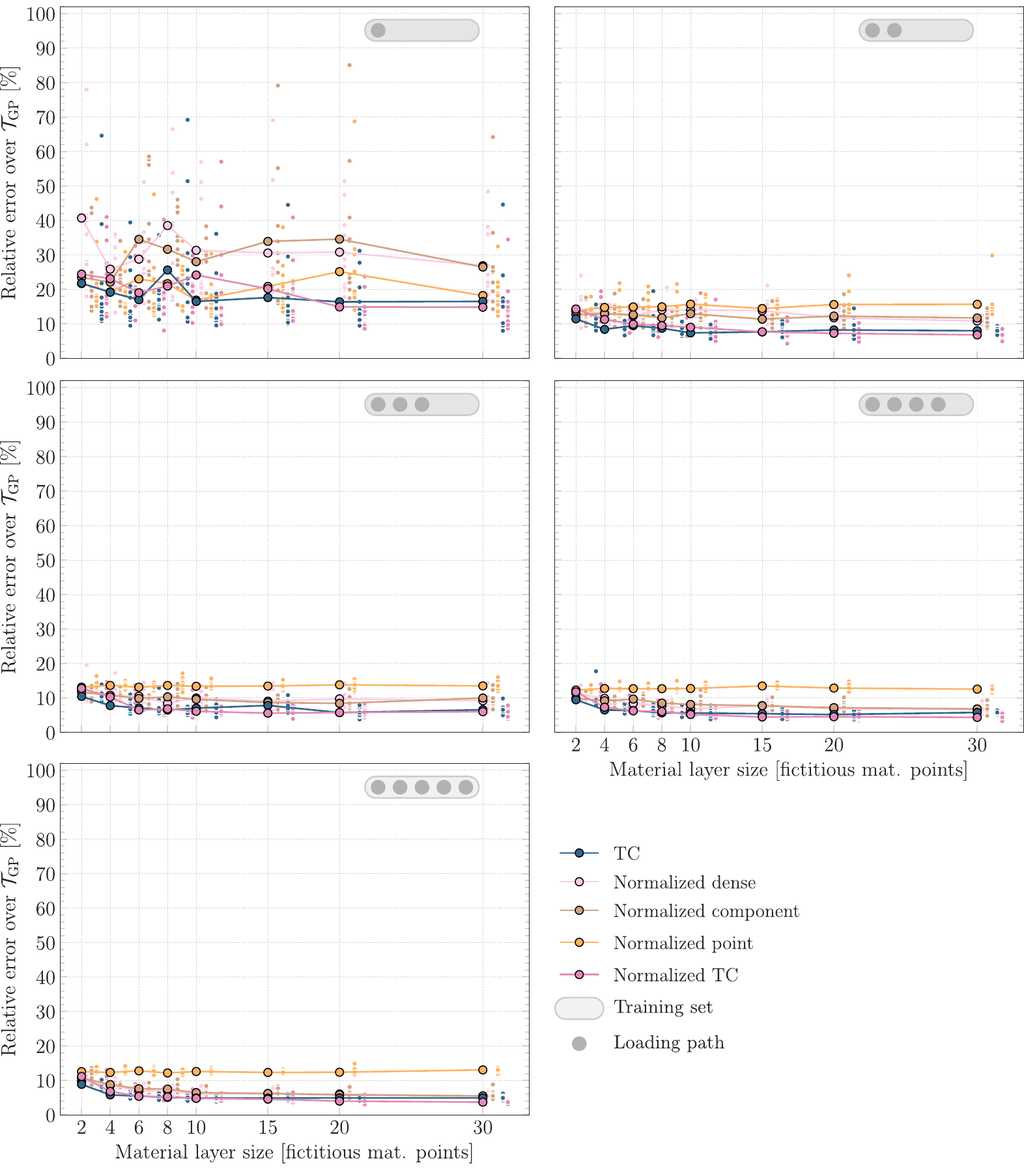}
\caption{Relative error over $\testunlgp$ for different combinations of material layer and training set size considering normalized decoders.}
\label{fig:normsvstctr01totr05}
\end{figure}

Having established in the previous section that the decoder connectivity design can improve generalization the low-data regime, we now investigate the impact of the weight constraint discussed in \Cref{sec:decoder} in improving training robustness and reducing even further training requirements. \Cref{fig:normsvstctr01totr05} shows the relative errors over the test set $\testunlgp$ for different training set and material layer sizes for all the normalized decoders and the \TCdec \ for reference, as it was the best performing type in \Cref{subsec:unnormaccuracy}, with solid lines referring to the average performance of each setup. 

A general remark is that the normalized decoders generally show superior performance to the unnormalized counterparts with the exception of the \TCdec, which remains the best decoder in terms of accuracy along with \NormTCdec. The normalized decoders show no degrading accuracy as more material points are considered and exhibit good generalization properties (\textit{i.e.} errors below 10\%) with training sets as small as two curves. Normalization also helps reduce the error variance. When training with small datasets, the randomly selected training paths might cover distinct parts of the design space or wander around a narrow portion of it. With large decoder weights, the model becomes overly sensitive to these variations. The constraint on the sum of the weights mitigates this effect by limiting the influence of any single material point, promoting more stable and consistent outputs across different training sets. 

\begin{figure}[!ht]
\centering
\includegraphics[width=.85\textwidth]{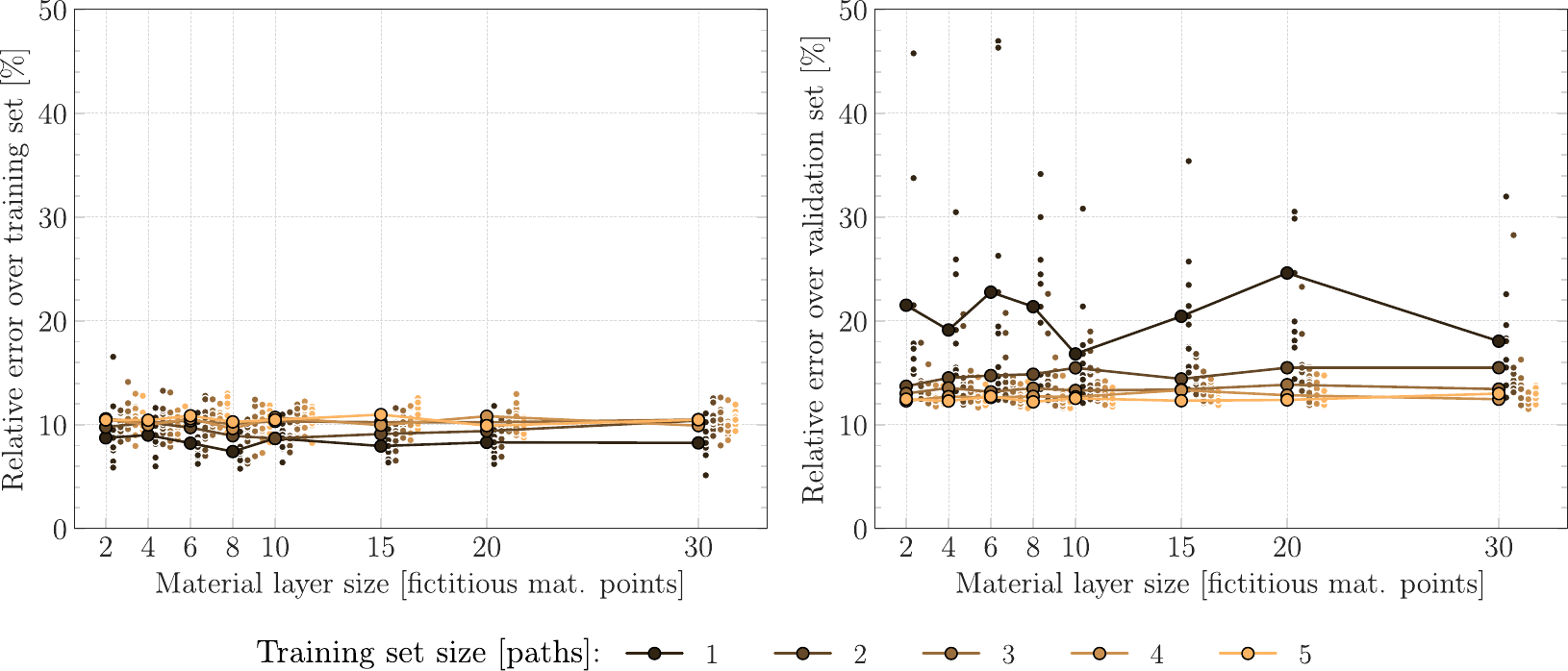}
\caption{Relative error over training and validation sets for PRNNs using the \Normpointdec \ decoder.}
\label{fig:pointstr01andtr05}
\end{figure}

With a very strong inductive bias, the \Normpointdec \ networks, in particular, reach errors as low as 13\% with a training set consisting of a single curve, proving to be, on average, a good option in this extreme scenario. However, as more training data are considered, the gain in accuracy is small and reaches a plateau around 3-4 curves with 12\% relative error, while the remaining options surpass it, with errors below 5\%. To understand why normalization is less effective in this architecture, we plot in \Cref{fig:pointstr01andtr05} the relative training and validation errors for a growing number of training curves. Although validation errors decrease when going from 1 to 2 training paths, they quickly reach a plateau around 3 curves, with errors around \qty{12}{\percent}. On the other hand, the error over the training set increases slightly as the training set grows, but also reaches a plateau around 10\%, with little effect from the number of material points. Therefore, despite relatively strong performance with only one training curve, weight normalization ends up overconstraining the architecture, leading to underfitting. 

Finally, while the normalization alone already greatly improves accuracy, reduces variance and training requirements, it does not enforce energy consistency between scales. The adjoint encoder-decoder structure combined with the scaling associated to each material point consistently provides the best overall performances across virtually all combinations of training set and material layer sizes, with or without normalization. Quite remarkably, both both \TCdec \ and \NormTCdec \ networks achieve errors below \qty{10}{\percent} having trained on a single curve. 

\section{Comparison of latent space variables with microscopic fields}
\label{sec:latent}

Previously, the effect of the different decoders was discussed in terms of accuracy of the homogenized stresses. In this section, we discuss and illustrate their impact on the latent space variables through a series of visualizations on representative (test) loading paths. Recall that our latent space is composed of strains, learned by the encoder, and stresses and internal variables, obtained from the embedded constitutive models. While the encoded strains could also be investigated, we focus here on the stresses and internal variables. All networks discussed in this section were trained on 5 curves, with the networks with the lowest relative test error being selected for the illustrations.

\subsection{Fictitious vs microscopic stresses}
\label{subsec:ficstresses}

Firstly, we compare the distribution of the fictitious stresses obtained in the material layer with the true full-field stress distribution of the micromodel for the different architectures explored in \Cref{sec:accuracylowdata}. We use two aggregate measures for that comparison: the mean and the maximum stress tensors, obtained by computing the mean and maximum separately for each stress component across the entire microscopic matrix field. Other metrics could be used to estimate the distance between the two distributions (true and fictitious). The Wasserstein distance, for example, provides more nuanced insights into the distribution differences, going beyond pointwise comparisons or aggregate errors. For this study, however, these two metrics already capture key aspects of the mismatch without introducing additional computational and interpretability burdens. 
\begin{figure}[!b]
\centering
\subfloat[Mean matrix stresses]{
\includegraphics[width=.9\textwidth]{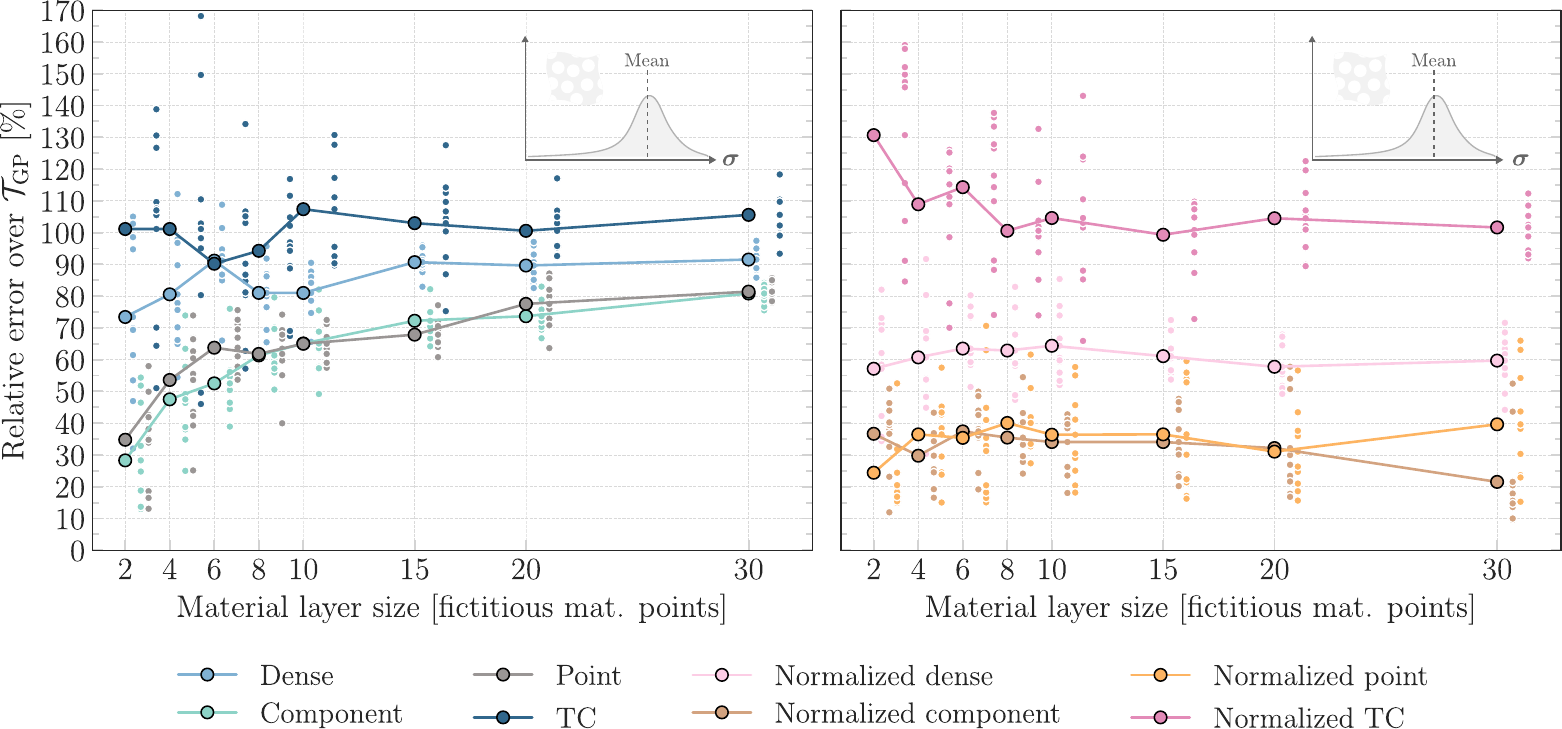}}
\hfill
\subfloat[Maximum matrix stresses]{
\includegraphics[width=.9\textwidth]{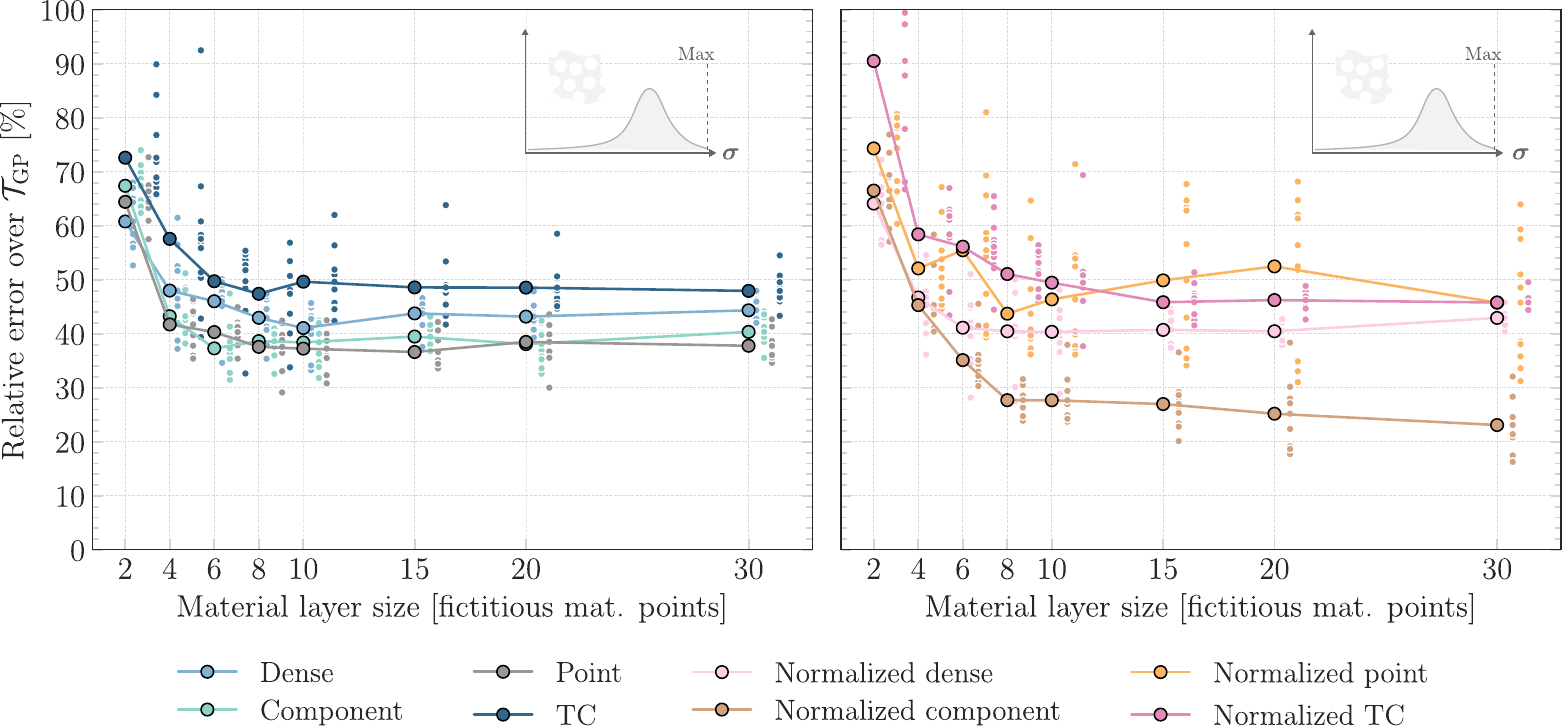}}
\caption{Relative error over $\testunlgp$ of fictitious stresses with different PRNN architectures trained on 5 paths compared to true microscopic stress distribution over matrix phase.}
\label{fig:errormeanmaxmatrix}
\end{figure}

\Cref{fig:errormeanmaxmatrix} shows the relative errors for the mean and maximum stresses over the test set $\testunlgp$ for the different decoders. Without normalization (left), all architectures struggle to capture either the mean or the maximum, stabilizing at high errors regardless of the number of material points. With normalization (right), errors in the mean stress remain high for the \Normdensedec \ and \NormTCdec, but reach lower levels for \Normcomponentdec \ and \Normpointdec, with errors as low as 10\%. For the maximum stress, only \Normcomponentdec \ shows a marked improvement, with errors below 20\% in the best-case scenario (larger networks), while the other alternatives remain comparable to the unnormalized decoders. 

It is worth noting that the networks with \TCdec \ and \NormTCdec, which were the best-performing in terms of accuracy of the homogenized stresses in \Cref{sec:accuracylowdata}, do not provide good agreement between the fictitious and microscopic stress distributions. There is, therefore, a trade-off between accuracy and interpretability. The additional constraints imposed by the self-adjoint architecture that improve the homogenized response do not result in fictitious stresses that are representative of the underlying microscopic fields. These results also indicate that normalization is crucial for aligning the fictitious distribution more closely with the true one, but it is not sufficient on its own. The best match between the distributions comes from \Normcomponentdec, an architecture that introduces some structure regarding cross-component stress contributions, unlike \Normdensedec, but still has enough flexibility to allow weights to vary per component, unlike \Normpointdec. 

The sparse architectures provide a useful alternative when interpretability of the latent space is sought in addition to accurate homogenization, while the \TCdec \ and \NormTCdec \ architectures are effective in obtaining accurate homogenized responses with extremely limited training data. The better microscopic alignment can be useful, for example, in extracting information from the latent space to predict microscopic quantities, which will be explored later in \Cref{sec:multitask}.

It is also worth mentioning that we started our comparison directly with the matrix phase only, we include in \ref{appendix:matrixorfullrve} a comparison of the similarity of the fictitious stresses with the full micromodel response (\textit{i.e.}, matrix + fibers). Based on the lowest error, the fictitious distributions are indeed closer to the matrix phase than to the full micromodel. This agreement is expected, given the dominance of matrix behavior in the current setup. However, for more intricate micromodels (\textit{e.g.}, a richer mix of constitutive behaviors and/or several phases), this correspondence may no longer hold and should be carefully examined. Nonetheless, the use of sparse connections and weight normalization is expected to remain beneficial, as these features promote robust training and more interpretable contributions regardless of the underlying micromodel complexity.
\begin{figure}[p]
\centering
\includegraphics[width=\textwidth]{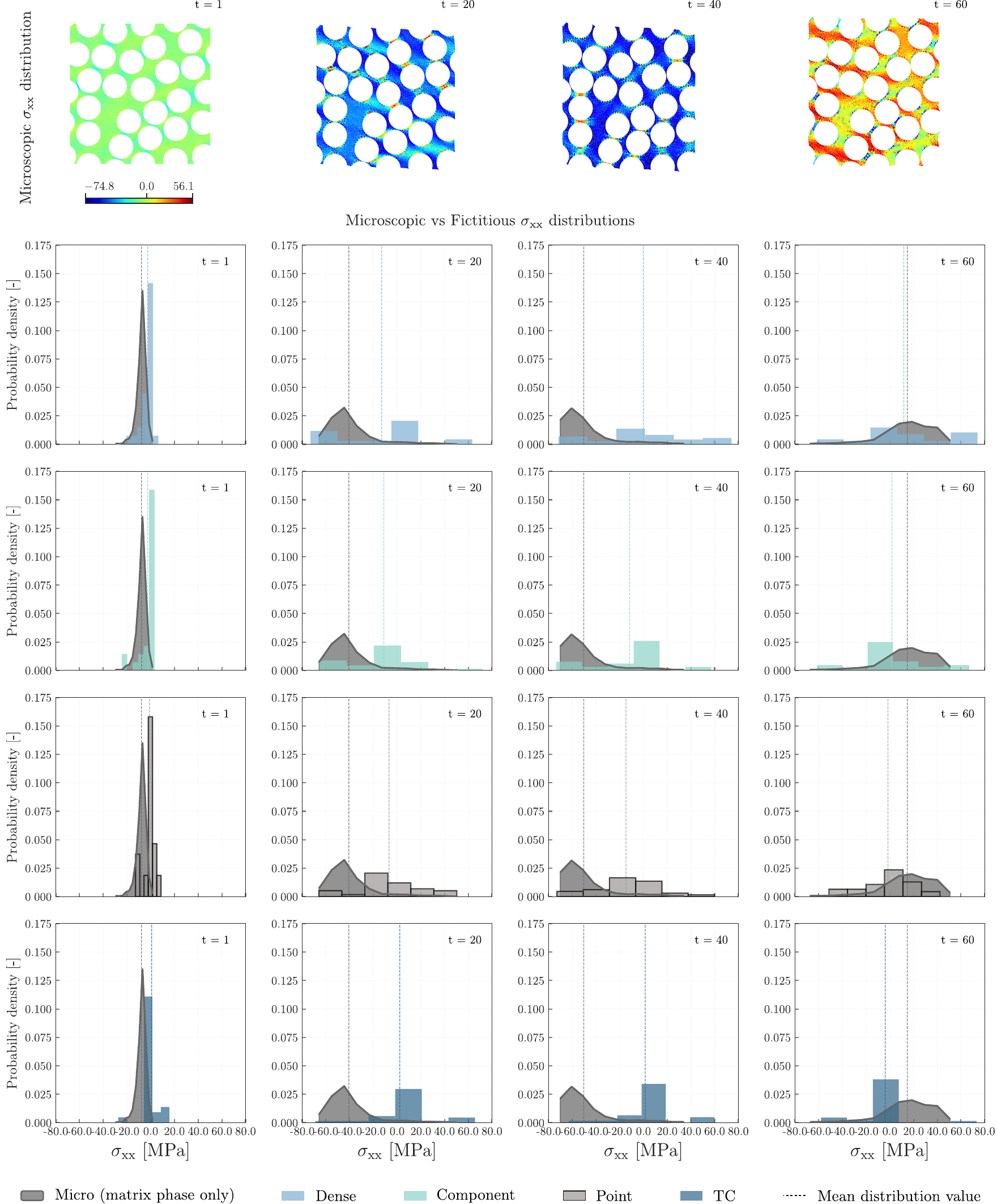}
\caption{Histogram distributions from the stresses computed in the matrix phase of the micromodel vs the stresses obtained from the fictitious material points of PRNNs with unnormalized decoders (color-coded) and 30 points at several time steps of an unseen GP-based path.}
\label{fig:histstresssamplegpxxunnorm}
\end{figure}

\begin{figure}[p]
\centering
\includegraphics[width=\textwidth]{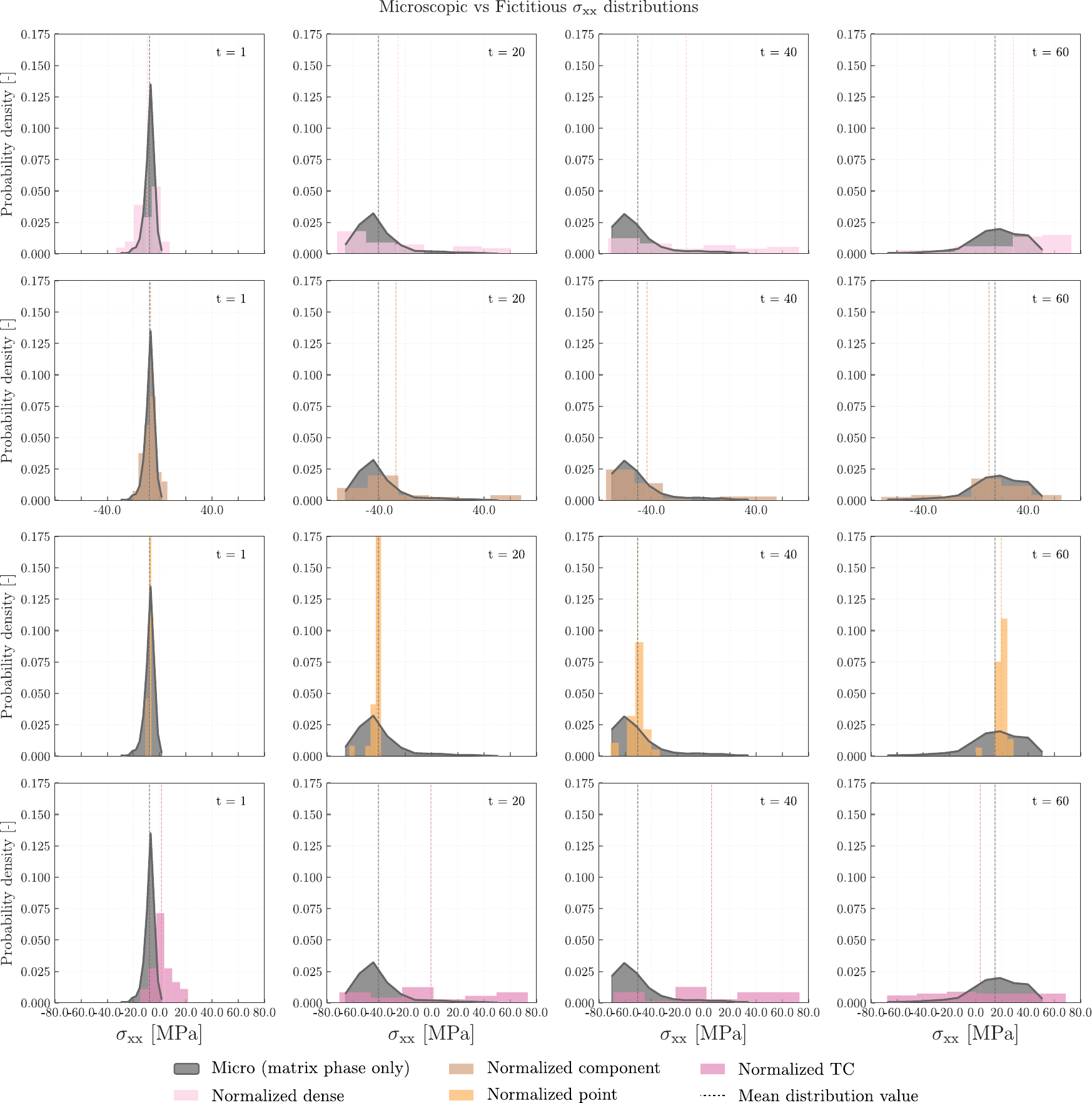}
\caption{Histogram distributions from the stresses computed in the matrix phase of the micromodel vs the stresses obtained from the fictitious material points of PRNNs with normalized decoders (color-coded) and 30 points at several time steps of an unseen GP-based path.}
\label{fig:histstresssamplegpxxnorm}
\end{figure}

Finally, \Cref{fig:histstresssamplegpxxunnorm} and \Cref{fig:histstresssamplegpxxnorm} show histograms of horizontal stresses at different time steps of an unseen GP-based curve, along with ground truth distributions coming from the micromodel. Each row corresponds to the snapshots of a specific type of decoder, color-coded accordingly. In general, only the distribution from the \Normcomponentdec \ shows a visually good match with the true stress distribution in the matrix, while the unnormalized alternatives show a more uniform distribution centered around zero across all time steps. \Normpointdec \ also captures the mean matrix stress quite well, but its accuracy in predicting the homogenized stress, our training target, is somewhat limited, as discussed in \Cref{subsec:weightnorm}. 

\subsection{Fictitious vs microscopic internal variables}
\label{subsec:intvar}

\begin{figure}[!ht]
\centering
\begin{minipage}{.75\textwidth}
\includegraphics[width=\textwidth]{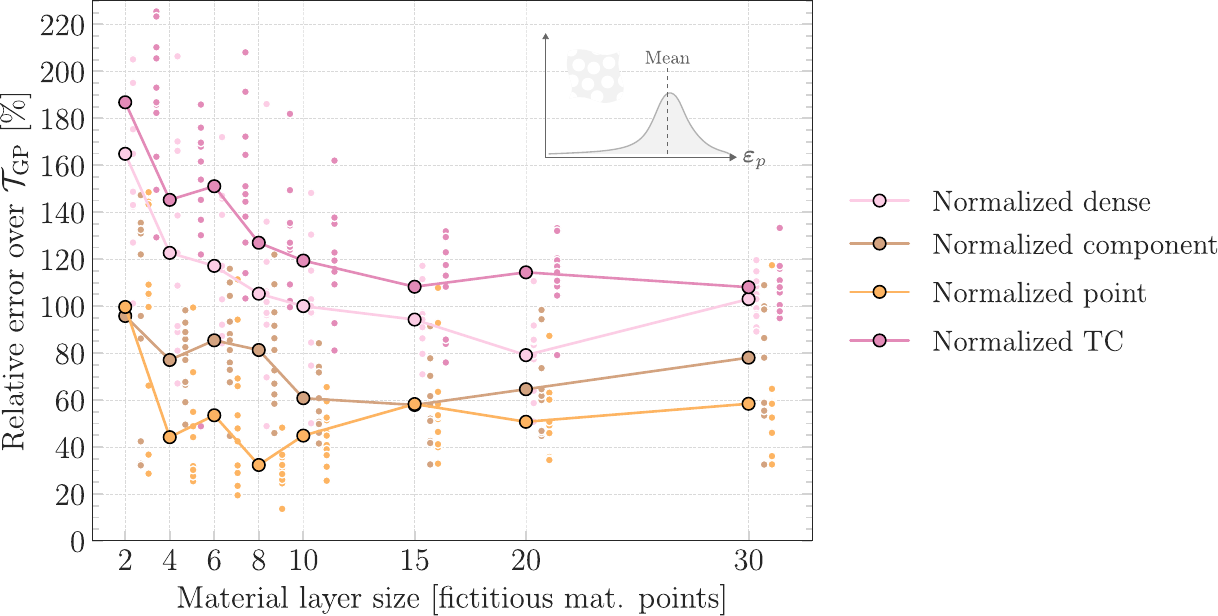} 
\end{minipage}
\caption{Relative error between mean fictitious plastic strains and mean microscopic plastic strains (matrix phase only) over paths from $\testunlgp$ using PRNNs with different decoders.}
\label{fig:errormicroepeq}
\end{figure}

Next, we investigate the similarity between the fictitious and the true internal variables of the micromodel. This time, we narrowed our study to the normalized decoders, as these showed overall better agreement in terms of microscopic stresses. \Cref{fig:errormicroepeq} shows that while the tailored changes regarding decoder sparsity help align the fictitious and true distributions, the average agreement remains significantly poorer than the stress comparison in \Cref{subsec:ficstresses}. This is expected since the networks are trained only on homogenized stress snapshots and only embed the matrix constitutive model, lacking the richer microstructural details present in the full RVE. For the \NormTCdec, we observe the same trend as in \Cref{fig:errormeanmaxmatrix}, with errors higher than the \Normdensedec, which is consistent with the trade-off between homogenization accuracy and latent space interpretability discussed previously. 

\section{Exploiting latent space interpretability for multitasking}
\label{sec:multitask}

Building on the findings in \Cref{sec:latent} regarding the similarity between the fictitious and the microscopic distributions, we now explore how this knowledge can be leveraged to predict not only homogenized quantities, but also relevant microscale information. In particular, we are interested in predicting the maximum hydrostatic stresses in the matrix, as these have been recently linked to macroscopic failure \cite{Clarijs2019,Martijnetal2023}. While the PRNN cannot predict failure directly, by establishing a threshold maximum stress one could, for instance, transition to a failure state (\textit{e.g.}, by inserting cohesive segments), switch to another surrogate model, or revert back to a full-order micromodel at the corresponding macroscopic integration point. To obtain the maximum hydrostatic stress in the matrix without modifying the current architecture, we add the following steps to the forward pass at each time step: 
\begin{enumerate}[label=\arabic*)]
\item with the stresses computed by material model, compute the hydrostatic stress for each fictitious material point;
\item from the pool of hydrostatic stresses, select the maximum.
\end{enumerate}

Our first test scenario employs the same networks considered in the previous sections - trained to predict $\widehat{\boldsymbol{\sigma}}^{\Omega}$ - to now predict our new quantity of interest $\sigma^{\omega}_{\textrm{hydro,max}}$. For this, we consider the test set $\testunlgp$ with 150 curves, and refer to the maximum hydrostatic stress values as $\testunlgp^{\textrm{max.}}$ when predicting the ``Maximum'' task (short for the maximum hydrostatic stress in the matrix) and $\testunlgp^{\textrm{avg.}}$ when predicting the ``Average'' task (short for the homogenized stresses). We focus on one specific architecture, the \Normcomponentdec, which showed the lowest error between the maximum fictitious stress and the maximum microscopic stress over the matrix phase in \Cref{subsec:ficstresses}. 

\begin{figure}[!h]
\centering
\includegraphics[width=.8\textwidth]{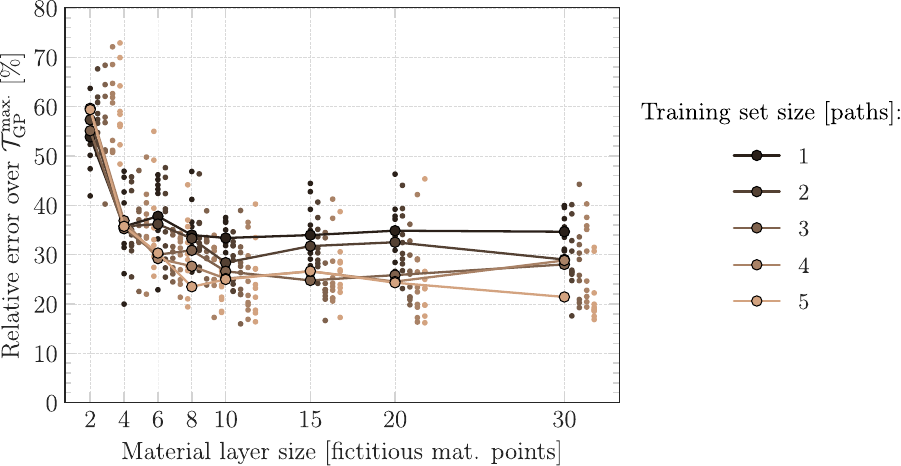}
\caption{Relative test set errors of \Normcomponentdec \ PRNNs over the unseen task of predicting the maximum hydrostatic matrix stress.}
\label{fig:unseenmax}
\end{figure}

The performance over the average task is discussed in detail in \Cref{sec:accuracylowdata}, where we show that with as few as two curves, relative errors in predicting the homogenized stresses remained consistently around 10\% or below across all material layer sizes tested. For the unseen Maximum task, \Cref{fig:unseenmax} shows that relative errors plateau around 30\% for training sets with two curves. With more training data (4-5 curves), the average error decreases to around 24\%, with the best performance at approximately 16\%. While these results show some correlation between the fictitious and true maximum hydrostatic stresses without training, the accuracy remains insufficient for a practical application. Therefore, in the next section, we exploit the interpretability of PRNNs to explicitly add the maximum hydrostatic matrix stress as a supervised target. 

\subsection{Incorporating maximum hydrostatic stress data}
\label{subsec:multidouble}

\begin{figure}[!ht]
\centering
\begin{minipage}{.9\textwidth}
\includegraphics[width=\textwidth]{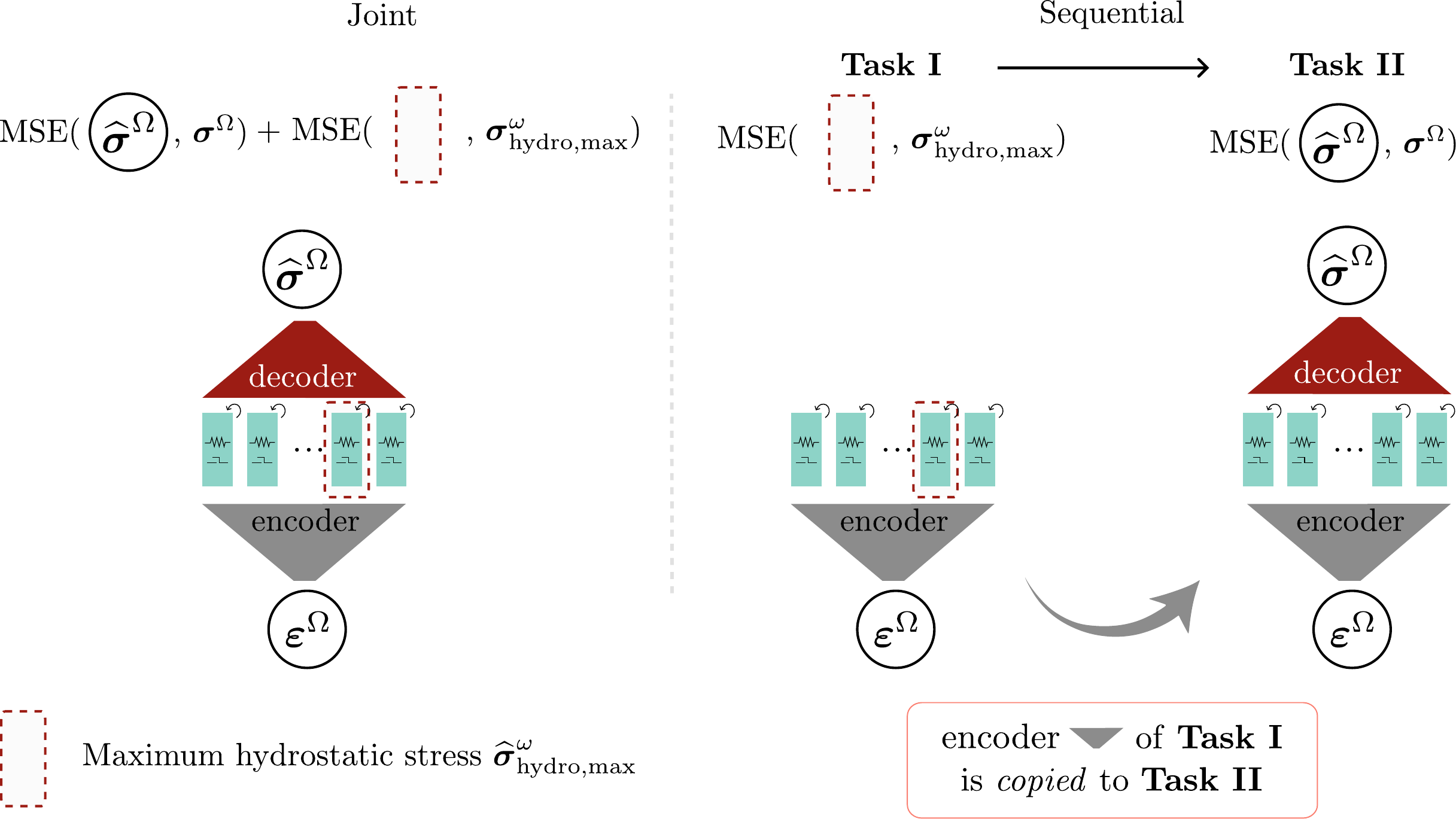}
\end{minipage}
\caption{Joint vs sequential learning for double-task model.}
\label{fig:doubleprnn}
\end{figure}

In this section, we investigate a double-task approach to predict the homogenized stress tensor and the maximum hydrostatic stress over the matrix phase. For that, two alternatives are investigated: a joint and a sequential learning approach. With joint learning, we compute an additional loss term related to the maximum task and add it to the term related to the average task as 
\begin{equation}
\label{eq:multi}
	\ell_{\textrm{double}} = \ell_{\textrm{avg.}} + \underbrace{\frac{1}{N} \ \sum_{i=1}^{N} \frac{1}{2} \ \| \ \sigma^{\omega}_{\textrm{hydro,max}}(\boldsymbol{\varepsilon}^{\Omega}_i) - \widehat{\sigma}^{\omega}_{\textrm{hydro,max}}(\boldsymbol{\varepsilon}^{\Omega}_i) \ \|^2}_{\ell_{\textrm{max.}}} 
\end{equation}
where $\ell_{\textrm{avg.}}$ is the loss function shown in \Cref{eq:lossavg}. In this case, encoder and decoder are trained with information from both tasks. In the sequential approach, encoder and decoder are instead trained separately. First, we train the encoder of the PRNN to predict the Maximum task. Since no trainable decoder is needed for that, in the second stage, the decoder is trained to learn the Average task while keeping the learned encoder from the Maximum task \textit{intact}. The two learning schemes are illustrated in \Cref{fig:doubleprnn}. It is important to note that for the sequential approach, it is not possible switching the order of the tasks since training for the Average task encompasses both encoder and decoder. 

\begin{figure}[!b]
\centering
\begin{minipage}[t]{.495\linewidth}
\subfloat[Errors over Maximum task\label{fig:jointseqmax}]{\includegraphics[width=\textwidth]{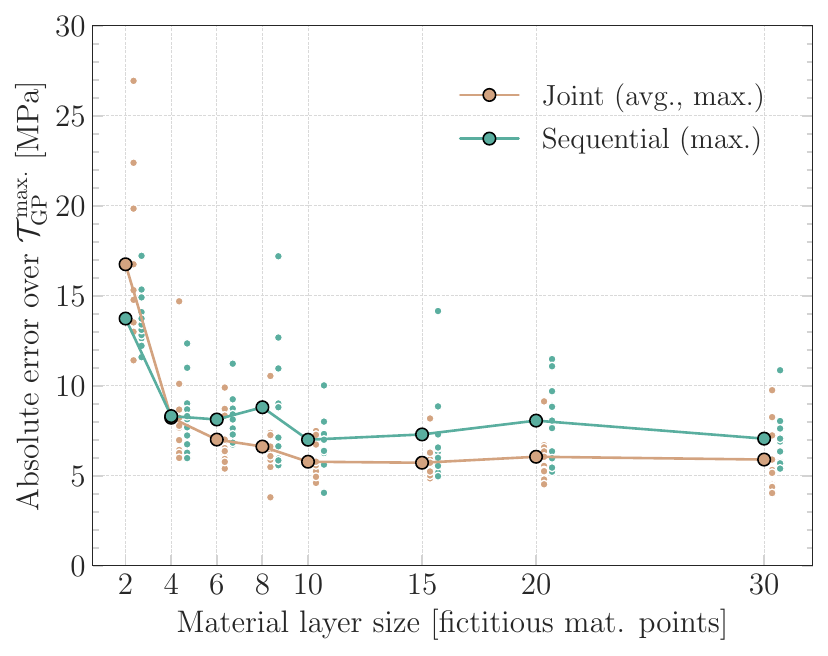}}
\end{minipage}
\begin{minipage}[t]{.495\linewidth}
\subfloat[Errors over Average task\label{fig:jointseqavg}]{\includegraphics[width=\textwidth]{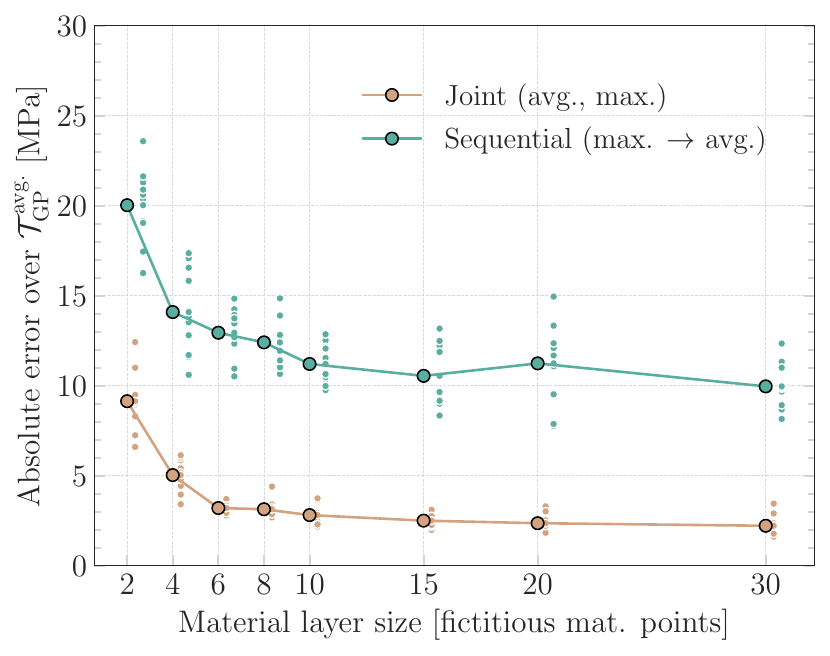}}
\end{minipage}
\label{fig:errorsjointvsseq}
\caption{Comparison of \textit{Normalized component} PRNNs trained on 5 curves adopting joint vs sequential learning on Maximum and Average tasks over test set $\testunlgp$.}
\end{figure}

\Cref{fig:jointseqmax} shows the errors of the two approaches on both tasks. On the Maximum task, there is only a slight gain in jointly learning the two tasks. However, in the second stage of the sequential approach, where the Average task is evaluated, the networks trained jointly show significantly lower errors. Recall that, in the second stage only the decoder is trainable, limiting the expressivity of the network - especially considering the constraints introduced by the sparsification and weight normalization. The joint learning approach is therefore the preferred approach considering both tasks. To further assess the effectiveness of this strategy, we compare its performance to that of training network independently for each task.

\subsection{Double-task vs single-task PRNNs}
\label{sub:multisingle}

In this section, we refer to the joint learning approach discussed in the previous section as ``Double'' and compare its performance with the two single-task counterparts ($\ell_{\textrm{max.}}$ and $\ell_{\textrm{avg.}}$) to assess the benefits and shortcomings of multitask learning. For this purpose, we use the networks trained to learn the Maximum task at the first stage of the sequential approach shown in \Cref{subsec:multidouble}, here referred to as Single (max.), while the networks trained to learn the homogenized stresses shown in \Cref{sec:accuracylowdata} are referred to as Single (avg.).

\begin{figure}[!t]
\begin{minipage}{.49\textwidth}
\subfloat[Average task\label{fig:avgtaskmultisingle}]{\includegraphics[width=\textwidth]{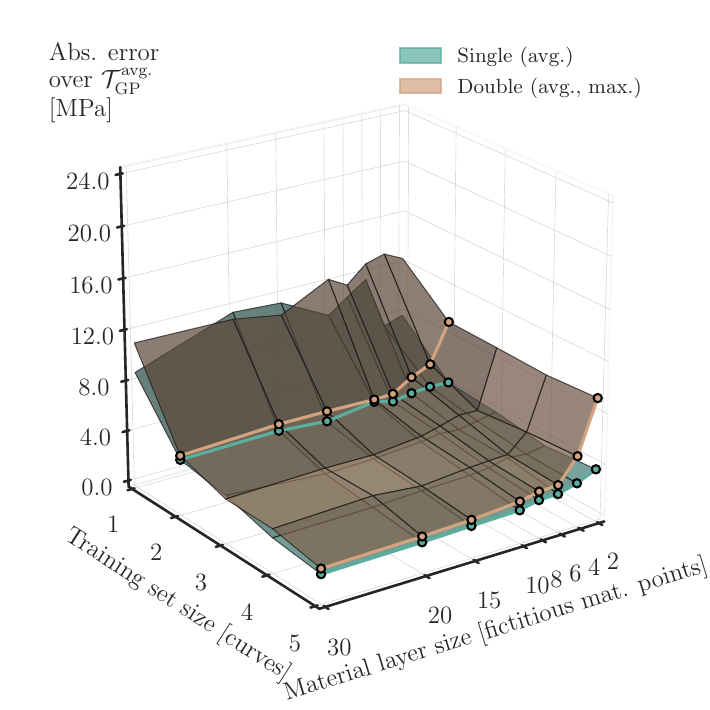}}
\end{minipage}
\begin{minipage}{.49\textwidth}
\subfloat[Maximum task\label{fig:maxtaskmultisingle}]{\includegraphics[width=\textwidth]{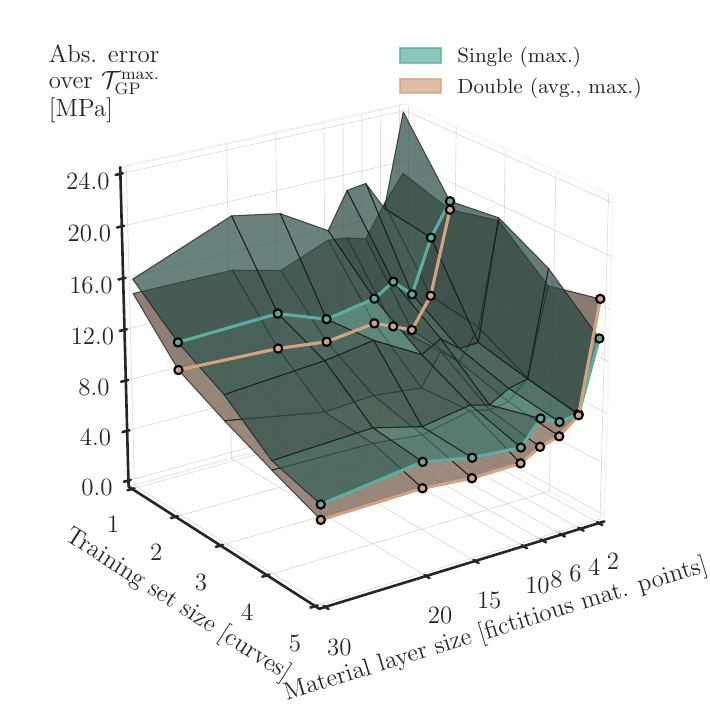}} 
\end{minipage}
\centering
\begin{minipage}{.9\textwidth}
\subfloat[Total error (Maximum + Average) of PRNNs trained on 2 and 5 GP-based curves\label{fig:multisingle2and5gps}]{\includegraphics[width=\textwidth]{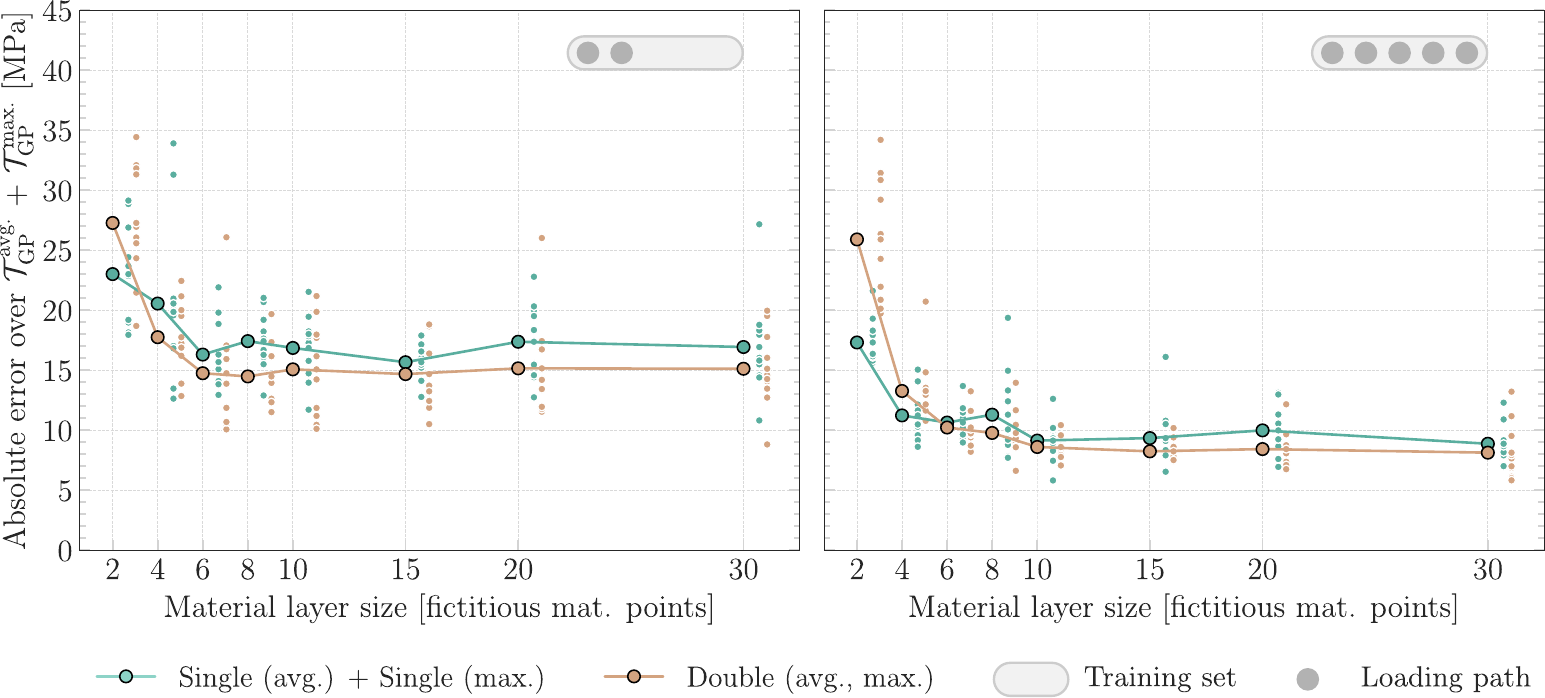}}
\end{minipage}
\caption{Test set errors of PRNNs trained for single and double-task and joint learning with limited data sets over the homogenized stress and maximum microscopic hydrostatic stress.}
\end{figure}

In \Cref{fig:avgtaskmultisingle}, we plot the 3D landscape of the errors for different training sets and material layer sizes on the Average task. For smaller networks (\textit{e.g.}, 2 and 4 material points), the difference between the single-task and double-task approaches is substantial across all training set sizes. For larger networks, this difference is minimal, and the two surfaces virtually overlap. Overall, jointly trained networks tend to trade off their accuracy on the Average task to capture the Maximum task better. Once the network is large enough to capture the complexity of both tasks, this trade-off vanishes, and the performance across both objectives becomes comparable to or better than that of the single-task counterparts.  

On the Maximum task (\Cref{fig:maxtaskmultisingle}), we observe the opposite trend. This time, the double-task approach yields a more accurate prediction, whereas the networks explicitly trained to predict the maximum hydrostatic stress exhibit slightly higher errors. This suggests that the Maximum task benefits from the additional information on the homogenized stresses, which is particularly useful for bringing the stresses from the latent space closer to the RVE stress distribution. The total error from the two single tasks is summed and shown in \Cref{fig:multisingle2and5gps} for two training set sizes, 2 and 5 curves. With 10 material points and 5 training curves, the total error for both tasks can go as low as 6 MPa. In terms of relative errors, this translates to a 7-10\% error in each task. 

\begin{figure}[!p]
\centering
\subfloat[Hydrostatic stresses from the fictitious material points of a PRNN\label{fig:hydrolatent}]{\includegraphics[width=.85\textwidth]{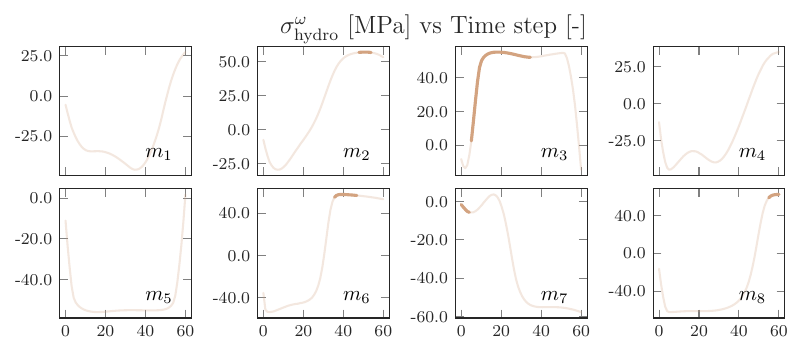}}
\hfill
\subfloat[Comparison with true maximum hydrostatic stresses  \label{fig:prnnmicrohydro}]{\includegraphics[width=.5\textwidth]{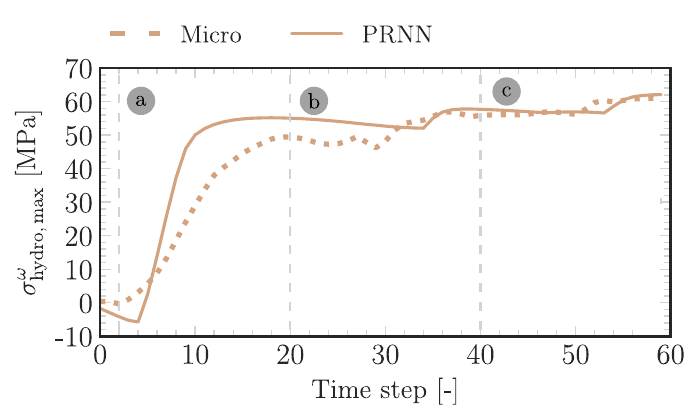}}
\hfill
\subfloat[Snapshots of microscopic distribution vs fictitious material points in PRNN \label{fig:prnnmicrosnapshots}]{\includegraphics[width=.825\textwidth]{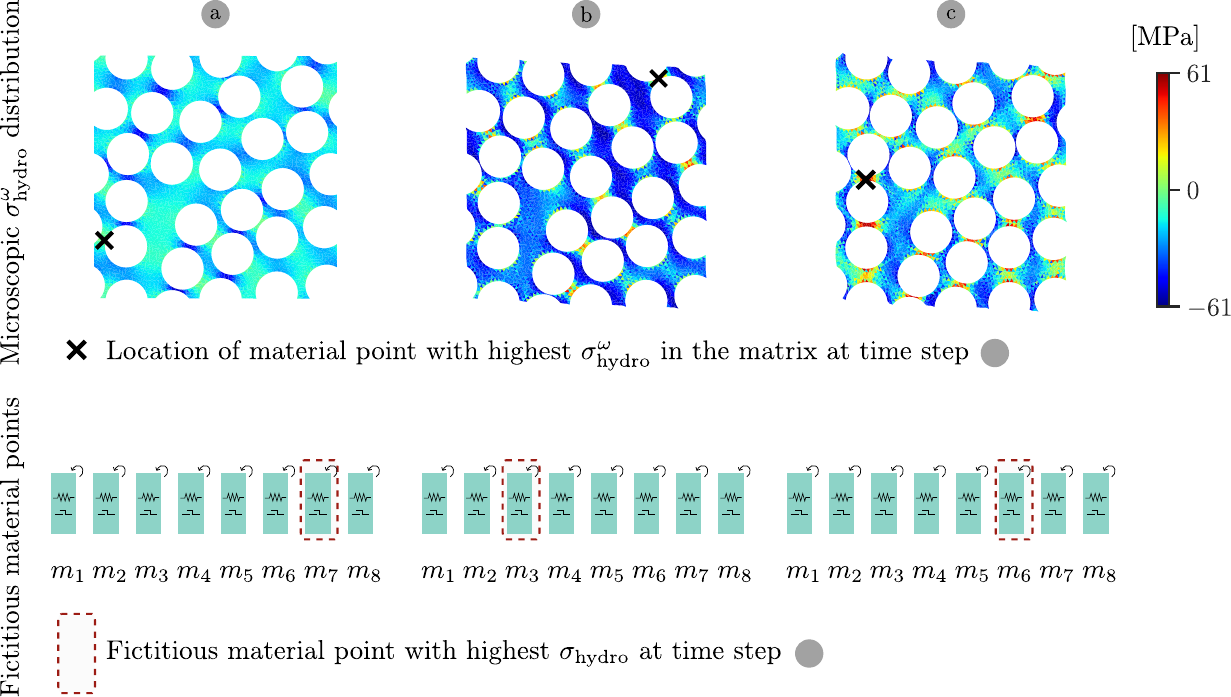}}
\caption{Hydrostatic stress from PRNN with 8 material points and trained on 5 GP-based curves on unseen loading path from $\testunlgp^{\textrm{max.}}$.}
\end{figure}

Next, we select the PRNN with the lowest error from the double-task approach to visualize the latent space of one loading path from the test set $\testunlgp$. In \Cref{fig:hydrolatent}, we show the hydrostatic stress evolution throughout 60 time steps for each one of the 8 fictitious material points and highlight only the parts where they are maximum. The combination of these parts is plotted in \Cref{fig:prnnmicrohydro} against the true maximum hydrostatic stress from the micromodel. Note that the fictitious material point responsible for the maximum changes throughout the loading path. This behavior is consistent with the evoling location of the maximum in the micromodel, as illustrated in \Cref{fig:prnnmicrosnapshots}, and helps explain why increasing the number of material points benefits the Maximum task, as larger material layers provide more candidates to represent the maximum hydrostatic stress.

\section{Final remarks}
\label{sec:conclusion}
In this work, we analysed different decoder architectures for PRNNs with linear encoders and decoders, focusing on how weight normalization, sparsity, and thermodynamic consistency affect accuracy and interpretability in the low-data regime. Based on the physical analogy with RVE volumes, we introduced a weight-normalization constraint that interprets decoder weights as relative volume contributions from fictitious material points. We showed that this constraint can improve not only the interpretability of the decoder, but also the robustness and accuracy of the network. We also introduced an adjoint encoder-decoder architecture with positive scalar contributions that provides a thermodynamically consistent formulation in terms of stress or energy, which enforces energetic consistency between the micro- and macroscopic scales and non-negative dissipation. With this architecture, as few as one training curve was sufficient to achieve homogenized stress prediction errors consistently below 10\%. Such data efficiency is particularly relevant when training on experimental or otherwise scarce data.

The effect of the normalization constraint, however, depends on the architecture it is applied to. In general, the constraint acts as regularization without additional hyperparameter tuning. It is effective in most cases, but it can also become overly restrictive (e.g., with the \Normpointdec), limiting the network's flexibility. Beyond its effect on the accuracy, normalization was also found to influence the latent representation. In particular, the (fictitious microscopic) stresses from the material layer can closely reflect their RVE counterparts without explicit training for this purpose. This correspondence has only been investigated for the material settings considered here and should therefore be interpreted with caution. Before exploiting such a correspondence in multi-task approaches, it should be verified for the specific material and architecture considered, as done in \Cref{sec:latent}. Nevertheless, this type of comparison provides insight into how directly the learned representation can reflect the underlying micromodel. It also highlights the trade-off between imposing additional physics-based structure and retaining interpretability in the latent representation. In this sense, sparse decoders facilitate this analysis by making the contributions of individual fictitious material points easier to disentangle and visualize.

Finally, the thermodynamically consistent formulation further demonstrates that architectural constraints can provide physical guarantees \textit{and} improve data efficiency. In our formulation, the network inherits the properties of the embedded material models and satisfies the Hill-Mandel principle of energy consistency between the microscopic and macroscopic scales. On the other hand, the fictitious and true microscopic distributions did not agree. This distinction points to the fact that thermodynamic consistency and correspondence with the microscopic state are separate properties of the architecture. Overall, the results demonstrate that decoder constraints, sparsity, and physically inspired architectural choices can be used not only to improve prediction accuracy and data efficiency, but also to retrieve and exploit the information in the latent space.

\biboptions{sort&compress}
\bibliographystyle{elsarticle-num-names} 
\bibliography{literature.bib}

@article{JI2026,
title = {Inelastic Constitutive Kolmogorov–Arnold Networks: A generalized framework for automated discovery of interpretable inelastic material models},
journal = {International Journal of Engineering Science},
volume = {228},
pages = {104630},
year = {2026},
issn = {0020-7225},
doi = {https://doi.org/10.1016/j.ijengsci.2026.104630},
url = {https://www.sciencedirect.com/science/article/pii/S0020722526001680},
author = {Chenyi Ji and Kian P. Abdolazizi and Hagen Holthusen and Christian J. Cyron and Kevin Linka}}

@article{SHIN2023,
  title = {Deep material network via a quilting strategy: visualization for explainability and recursive training for improved accuracy},
  volume = {9},
  ISSN = {2057-3960},
  DOI = {10.1038/s41524-023-01085-6},
  number = {1},
  journal = {npj Computational Materials},
  publisher = {Springer Science and Business Media LLC},
  author = {Shin,  Dongil and Alberdi,  Ryan and Lebensohn,  Ricardo A. and Dingreville,  Rémi},
  year = {2023},
  month = July 
}

@article{FLASCHEL2025,
title = {Convex neural networks learn generalized standard material models},
journal = {Journal of the Mechanics and Physics of Solids},
volume = {200},
pages = {106103},
year = {2025},
issn = {0022-5096},
doi = {https://doi.org/10.1016/j.jmps.2025.106103},
url = {https://www.sciencedirect.com/science/article/pii/S0022509625000791},
author = {Moritz Flaschel and Paul Steinmann and Laura {De Lorenzis} and Ellen Kuhl}
}

@article{RAJETAL2026,
title = {A neural-network framework to learn history-dependent constitutive laws and identifiability of internal variables},
journal = {Journal of the Mechanics and Physics of Solids},
volume = {217},
pages = {106807},
year = {2026},
issn = {0022-5096},
doi = {https://doi.org/10.1016/j.jmps.2026.106807},
author = {Mayank Raj and Lianghao Cao and Andrew Stuart and Kaushik Bhattacharya}
}

@article{CHEUNG2024b,
title = {Augmentation of scarce data — A new approach for deep-learning modeling of composites},
journal = {Composites Science and Technology},
volume = {249},
pages = {110491},
year = {2024},
issn = {0266-3538},
doi = {https://doi.org/10.1016/j.compscitech.2024.110491},
author = {Hon Lam Cheung and Petter Uvdal and Mohsen Mirkhalaf}}

@article{EGHBALIAN2023,
	author = {Mahdad Eghbalian and Mehdi Pouragha and Richard Wan},
	doi = {https://doi.org/10.1016/j.compgeo.2023.105472},
	issn = {0266-352X},
	journal = {Computers and Geotechnics},
	pages = {105472},
	title = {A physics-informed deep neural network for surrogate modeling in classical elasto-plasticity},
	volume = {159},
	year = {2023},
}

@inbook{Geers2017,
	author = {Geers, Marc G. D. and Kouznetsova, Varvara G. and Matouš, Karel and Yvonnet, Julien},
	booktitle = {Encyclopedia of Computational Mechanics Second Edition},
	chapter = {},
	doi = {https://doi.org/10.1002/9781119176817.ecm2107},
	eprint = {https://onlinelibrary.wiley.com/doi/pdf/10.1002/9781119176817.ecm2107},
	isbn = {9781119176817},
	pages = {1-34},
	publisher = {John Wiley \& Sons, Ltd},
	title = {Homogenization Methods and Multiscale Modeling: Nonlinear Problems},
	year = {2017},
}

@article{Koeppe2021, 
title={Explainable artificial intelligence for mechanics: Physics-Explaining neural networks for constitutive models}, 
volume={8}, 
DOI={10.3389/fmats.2021.824958}, journal={Frontiers in Materials},
author={Koeppe, Arnd and Bamer, Franz and Selzer, Michael and Nestler, Britta and Markert, Bernd}, 
year={2022}}

@article{kovacs2024,
title = {Physically Recurrent Neural Networks for computational homogenization of composite materials with microscale debonding},
journal = {European Journal of Mechanics - A/Solids},
volume = {112},
pages = {105668},
year = {2025},
issn = {0997-7538},
doi = {https://doi.org/10.1016/j.euromechsol.2025.105668},
author = {N. Kovács and M.A. Maia and I. B. C. M. Rocha and C. Furtado and P. P. Camanho and F. P. {van der Meer}}}

@misc{kovacs2025,
      title={Uncertainty Quantification in Multiscale Modeling of Polymer Composite Materials Using Physically Recurrent Neural Networks}, 
      author={N. Kovács and I. B. C. M. Rocha and F. P. van der Meer and C. Furtado and P. P. Camanho},
      year={2025},
      eprint={2504.11625},
      archivePrefix={arXiv},
      primaryClass={cond-mat.dis-nn},
      url={https://arxiv.org/abs/2504.11625}, 
      shorthand = {**} 
}

@article{HE2022,
title = {Thermodynamically consistent machine-learned internal state variable approach for data-driven modeling of path-dependent materials},
journal = {Computer Methods in Applied Mechanics and Engineering},
volume = {402},
pages = {115348},
year = {2022},
note = {A Special Issue in Honor of the Lifetime Achievements of J. Tinsley Oden},
issn = {0045-7825},
doi = {https://doi.org/10.1016/j.cma.2022.115348},
author = {Xiaolong He and Jiun-Shyan Chen}
}

@article{Maiaetal2023,
	author = {M. A. Maia and Rocha, I. B. C. M. and P. Kerfriden  and F. P. van der Meer},
	doi = {https://doi.org/10.1016/j.cma.2023.115934},
	journal = {Computer Methods in Applied Mechanics and Engineering},
	pages = {115934},
	title = {Physically recurrent neural networks for path-dependent heterogeneous materials: Embedding constitutive models in a data-driven surrogate},
	volume = {407},
	year = {2023}}

@article{Maiaetal2024,
title = {Physically recurrent neural network for rate and path-dependent heterogeneous materials in a finite strain framework},
journal = {Mechanics of Materials},
volume = {198},
pages = {105145},
year = {2024},
doi = {https://doi.org/10.1016/j.mechmat.2024.105145},
author = {M. A. Maia and I. B. C. M. Rocha and D. Kovačević and F. P. van der Meer}}

@misc{ghane2025woven,
      title={Multiscale Analysis of Woven Composites Using Hierarchical Physically Recurrent Neural Networks},
      author={Ehsan Ghane and M. A. Maia and Iuri B. C. M. Rocha and Martin Fagerström and Mohsen Mirakhalaf},
      year={2025},
      eprint={2503.04901},
      archivePrefix={arXiv},
      primaryClass={physics.comp-ph},
      url={https://arxiv.org/abs/2503.04901},
}

@article{MASI2026,
title = {Learning inelastic constitutive models from stress–strain data under hard thermodynamic constraints},
journal = {Computer Methods in Applied Mechanics and Engineering},
volume = {461},
pages = {119260},
year = {2026},
issn = {0045-7825},
doi = {https://doi.org/10.1016/j.cma.2026.119260},
url = {https://www.sciencedirect.com/science/article/pii/S0045782526005335},
author = {Filippo Masi}}

@article{Rosenkranz2023,
author = {Rosenkranz, Max and Kalina, Karl A. and Brummund, Jörg and Kästner, Markus},
title = {A comparative study on different neural network architectures to model inelasticity},
journal = {International Journal for Numerical Methods in Engineering},
volume = {124},
number = {21},
pages = {4802-4840},
doi = {https://doi.org/10.1002/nme.7319},
year = {2023}}

@article{Mozaffar26414,
	author = {Mozaffar, M. and Bostanabad, R. and Chen, W. and Ehmann, K. and Cao, J. and Bessa, M. A.},
	doi = {10.1073/pnas.1911815116},
	issn = {0027-8424},
	journal = {Proceedings of the National Academy of Sciences},
	number = {52},
	pages = {26414--26420},
	publisher = {National Academy of Sciences},
	title = {Deep learning predicts path-dependent plasticity},
	volume = {116},
	year = {2019},
}

@article{VLASSIS2021,
	author = {Vlassis, N. N. and Sun, WaiChing},
	doi = {https://doi.org/10.1016/j.cma.2021.113695},
	issn = {0045-7825},
	journal = {Computer Methods in Applied Mechanics and Engineering},
	pages = {113695},
	title = {Sobolev training of thermodynamic-informed neural networks for interpretable elasto-plasticity models with level set hardening},
	volume = {377},
	year = {2021},
}

@phdthesis{Clarijs2019,
title = "Mechanical performance of glassy polymers: influence of physical ageing and molecular architecture",
author = "Coen Clarijs",
year = "2019",
month = may,
day = "10",
isbn = "978-90-386-4741-8",
publisher = "Technische Universiteit Eindhoven",
type = "PhD thesis",
school = "Mechanical Engineering",
}

@article{Martijnetal2023,
author = {Wismans, Martijn and van den Broek, Sten J. J. and van Breemen, Lambèrt C. A. and Govaert, Leon E. and Engels, Tom A. P.},
title = {Micromechanical analysis of age-induced strength reduction in a multiaxially loaded short-fiber reinforced thermoplastic},
journal = {Journal of Applied Polymer Science},
volume = {140},
number = {41},
pages = {e54512},
doi = {https://doi.org/10.1002/app.54512},
year = {2023}}

@article{GHANE2025,
title = {Multi-fidelity data fusion for inelastic woven composites: Combining recurrent neural networks with transfer learning},
journal = {Composites Science and Technology},
volume = {267},
pages = {111163},
year = {2025},
issn = {0266-3538},
doi = {https://doi.org/10.1016/j.compscitech.2025.111163},
author = {Ehsan Ghane and Martin Fagerström and Mohsen Mirkhalaf}
}

@article{Guo2021cpi,
  title = {CPINet: Parameter identification of path-dependent constitutive model with automatic denoising based on CNN-LSTM},
  volume = {90},
  ISSN = {0997-7538},
  url = {http://dx.doi.org/10.1016/j.euromechsol.2021.104327},
  DOI = {10.1016/j.euromechsol.2021.104327},
  journal = {European Journal of Mechanics - A/Solids},
  publisher = {Elsevier BV},
  author = {Guo,  Zhenfei and Bai,  Ruixiang and Lei,  Zhenkun and Jiang,  Hao and Liu,  Da and Zou,  Jianchao and Yan,  Cheng},
  year = {2021},
  month = Nov,
  pages = {104327}
}

@article{HEIDENREICH2025,
title = {Extended minimal state cells (EMSC): Self-consistent recurrent neural networks for rate- and temperature dependent plasticity},
journal = {International Journal of Plasticity},
volume = {188},
pages = {104305},
year = {2025},
issn = {0749-6419},
doi = {https://doi.org/10.1016/j.ijplas.2025.104305},
url = {https://www.sciencedirect.com/science/article/pii/S0749641925000646},
author = {Julian N. Heidenreich and Dirk Mohr}
}

@article{BonattiMohr2021,
author = {Colin Bonatti  and Dirk Mohr },
title = {One for all: Universal material model based on minimal state-space neural networks},
journal = {Science Advances},
volume = {7},
number = {26},
pages = {eabf3658},
year = {2021},
doi = {10.1126/sciadv.abf3658},
URL = {https://www.science.org/doi/abs/10.1126/sciadv.abf3658}}

\appendix
\section{Matrix vs full micromodel}
\label{appendix:matrixorfullrve}
\begin{figure}[!b]
\centering
\subfloat[Normalized component - Mean stresses\label{fig:meanmatrixfullsparsenorm}]{\includegraphics[width=.49\textwidth]{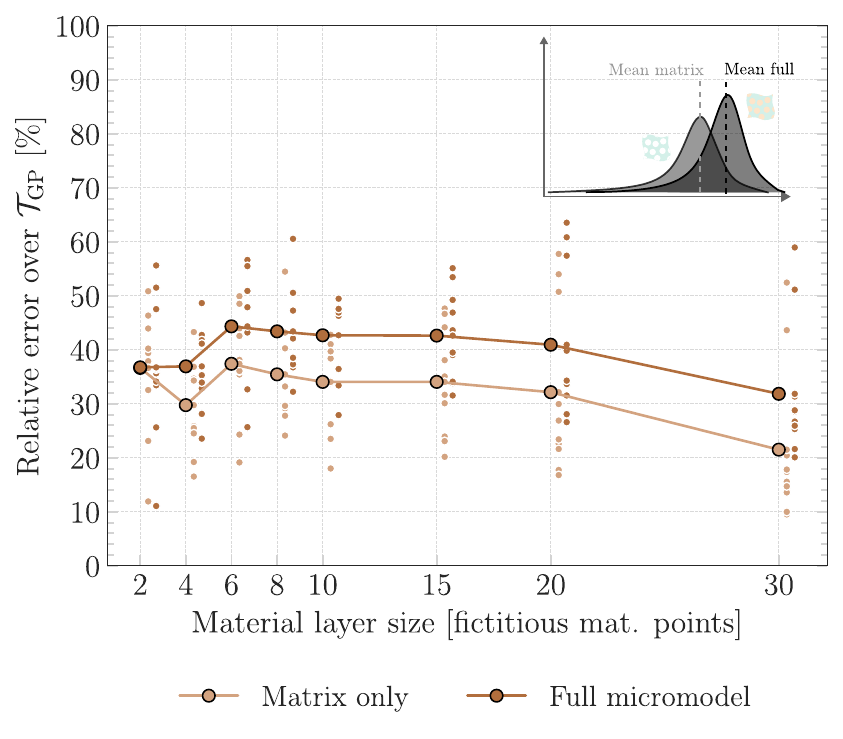}}
\hfill
\subfloat[Normalized component - Maximum stresses\label{fig:unseensparsenormmax}]{\includegraphics[width=.49\textwidth]{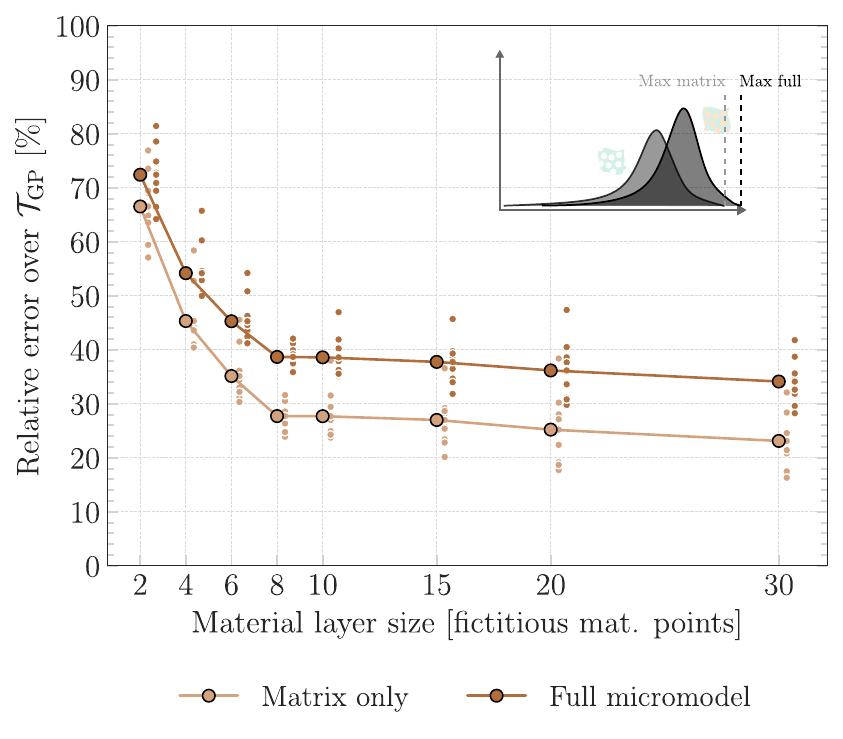}}
\hfill
\subfloat[Normalized point - Mean stresses\label{fig:meanmatrixfullpointnorm}]{\includegraphics[width=.49\textwidth]{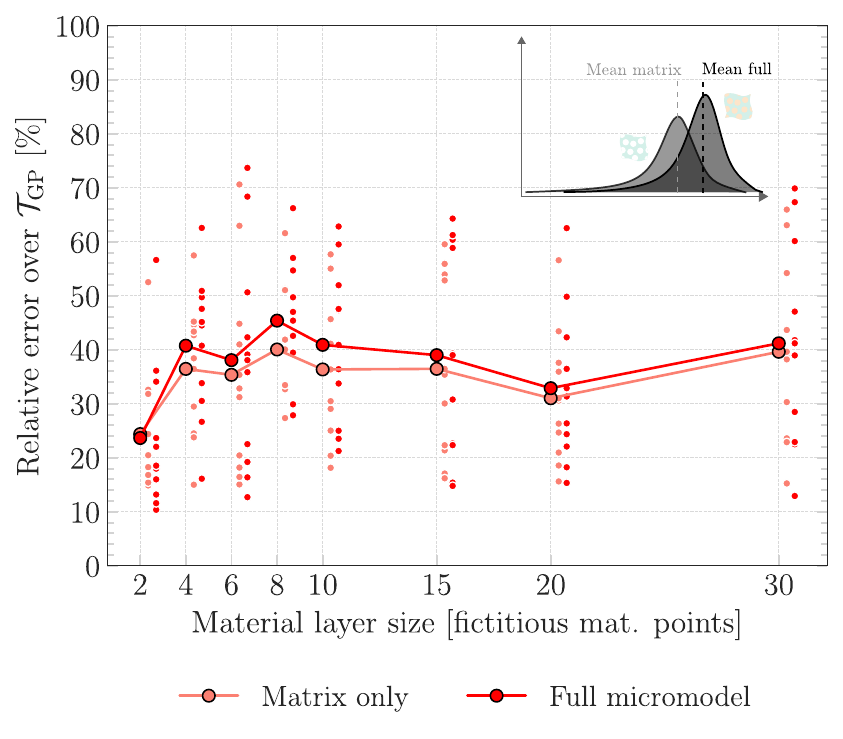}}
\hfill
\subfloat[Normalized point - Maximum stresses\label{fig:unseenpointnormmax}]{\includegraphics[width=.49\textwidth]{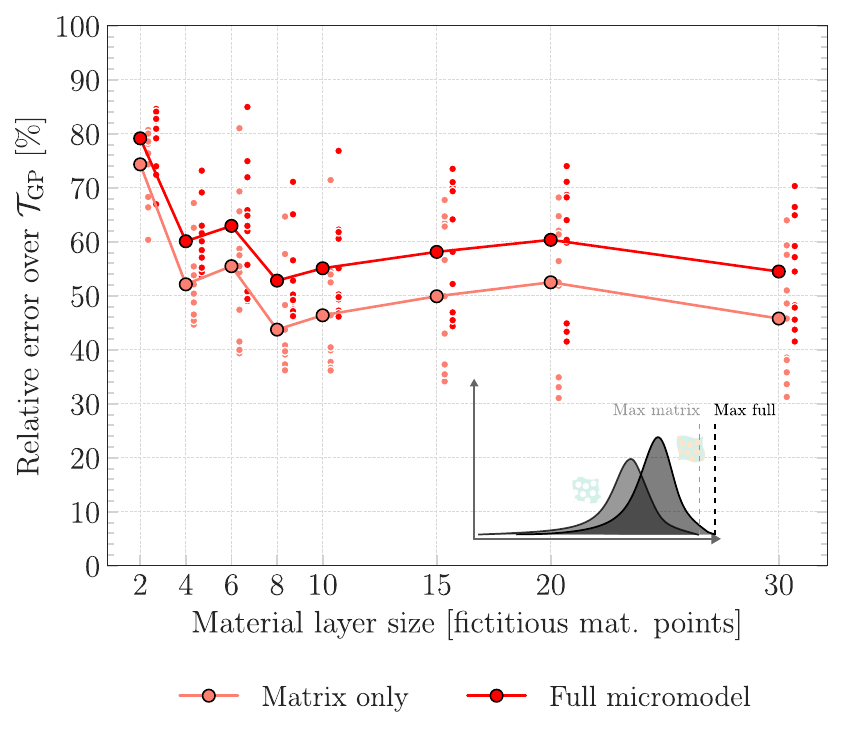}}
\hfill \strut
\caption{Relative error of fictitious stresses from normalized and sparse decoders trained on 5 curves compared to true microscopic stresses over matrix phase and full micromodel.}
\label{fig:matrixorfullrve}
\end{figure}

\Cref{fig:matrixorfullrve} shows the relative error between the mean and maximum stresses from the fictitious material points of PRNNs with different decoders trained on 5 curves and the true microscopic stresses over two domains: matrix phase only and full micromodel. For both metrics, the difference between the fictitious and the true stresses grows as the networks gains complexity and plateaus around 10\%, with the match with the matrix phase quantities being the best overall.

\end{document}